\documentclass[letterpaper, 10 pt, conference]{ieeeconf}  % Comment this line out if you need a4paper
\usepackage{times}
\usepackage{epsfig}
\usepackage{graphicx}
\usepackage{amsmath}
\usepackage{amssymb}
\usepackage{booktabs}
\usepackage{verbatim}
\usepackage{algorithm}  
\usepackage{algpseudocode}  
\usepackage{amsmath} 
\usepackage{threeparttable}

\IEEEoverridecommandlockouts                              % This command is only needed if 
\makeatletter

\newcommand{\Rmnum}[1]{\expandafter\@slowromancap\romannumeral #1@}
\makeatother
\title{\LARGE \bf
Lightweight Object-level Topological Semantic Mapping and Long-term Global Localization based on Graph Matching
}

\author{Fan Wang$^{1}$$^{2}$, Chaofan Zhang$^{1}$$^{*}$, Fulin Tang$^{3}$, Hongkui Jiang$^{1}$$^{2}$, Yihong Wu$^{3}$, and Yong Liu$^{1}$% <-this % stops a space
\thanks{This work was supported by The Key Research and Development program of Anhui Province of China(202104a05020043); National Natural Science Foundation of China(62102395); the Open Projects Program of National Laboratory of Pattern Recognition(20210040); Natural Science Foundation of Anhui Province of China(2108085QF277).}% <-this % stops a space
\thanks{$^{1}$Anhui Institute of Optics and Fine Mechanics, and Hefei Institutes of Physical Science, Chinese Academy of Sciences, Hefei 230031, P. R. China.
        {\tt\small (e-mail: zcfan@aiofm.ac.cn; liuyong@aiofm.ac.cn)}}%
\thanks{$^{2}$University of Science and Technology of China, Hefei 230026, P. R. China.
	{\tt\small (e-mail: wangfan8@mail.ustc.edu.cn; jhkjhk@mail.ustc.edu.cn) }}%
\thanks{$^{3}$National Laboratory of Pattern Recognition, Institute of Automation, Chinese Academy of Sciences, Beijing 100864, China.
	{\tt\small (e-mail:  yihong.wu@ia.ac.cn; fulin.tang@nlpr.ia.ac.cn) }}%
\thanks{$^{*}$is the corresponding author. }
\thanks{Fan Wang and Chaofan Zhang contributed equally to this work.}
}

\begin{document}

\maketitle
\thispagestyle{empty}
\pagestyle{empty}

%%%%%%%%%%%%%%%%%%%%%%%%%%%%%%%%%%%%%%%%%%%%%%%%%%%%%%%%%%%%%%%%%%%%%%%%%%%%%%%%
\begin{abstract}
Mapping and localization are two essential tasks for mobile robots in real-world applications.
However, large-scale and dynamic scenes challenge the accuracy and robustness of most current mature solutions. 
This situation becomes even worse when computational resources are limited.
In this paper, we present a novel lightweight object-level mapping and localization method with high accuracy and robustness.
Different from previous methods, our method does not need a prior constructed precise geometric map, which greatly releases the storage burden, especially for large-scale navigation.
% Unlike previous methods that construct precise geometric map first, we only xxx, which .
We use object-level features with both semantic and geometric information to model landmarks in the environment.
% We propose a novel approach to construct descriptors while increasing their discriminability and match-ability, based on recent advances in neural networks.
Particularly, a learning topological primitive is first proposed to efficiently obtain and organize the object-level landmarks.
On the basis of this, we use a robot-centric mapping framework to represent the environment as a semantic topology graph and relax the burden of maintaining global consistency at the same time.
Besides, a hierarchical memory management mechanism is introduced to improve the efficiency of online mapping with limited computational resources.
% Based on the proposed map, a robust localization method that enhances the descriptors of nodes by constructing local semantic scene graphs, and performing robust localization by comparing scene similarity based on graph matching.
Based on the proposed map, the robust localization is achieved by constructing a novel local semantic scene graph descriptor, and performing multi-constraint graph matching to compare scene similarity.
Finally, we test our method on a low-cost embedded platform to demonstrate its advantages.
Experimental results on a large scale and multi-session real-world environment show that the proposed method outperforms the state of arts in terms of lightweight and robustness.
% Additionally, these systems usually face further challenges, such as limited computational power, or insufficient memory for storing large maps of the entire environment.
% Thus, developing lightweight map representations is of considerable interest for enabling large-scale xx.
\end{abstract}

%%%%%%%%%%%%%%%%%%%%%%%%%%%%%%%%%%%%%%%%%%%%%%%%%%%%%%%%%%%%%%%%%%%%%%%%%%%%%%%%
\section{Introduction}

Mobile robots have gained impressive developments in various fields over the past decades.
Robust mapping and localization are critical prerequisites for the long-term autonomous navigation of mobile robots.
The most important solution is to establish an effective global consistency description of the environmental information, and simultaneous localization~\cite{c1}. 
%Classical  global consistent description and localization methods using geometric and visual feature have achieved good results and promoted the application and deployment of robots. 
%However, these methods still face challenges for long-term navigation in large, unstructured and dynamic environments.
Classical metric maps build high precise geometric model of environment and clearly correspond to the real world, which based on SLAM technology.
%\begin{comment}
\begin{figure}[htbp]
	\begin{center}
		\includegraphics[width=1 \linewidth]{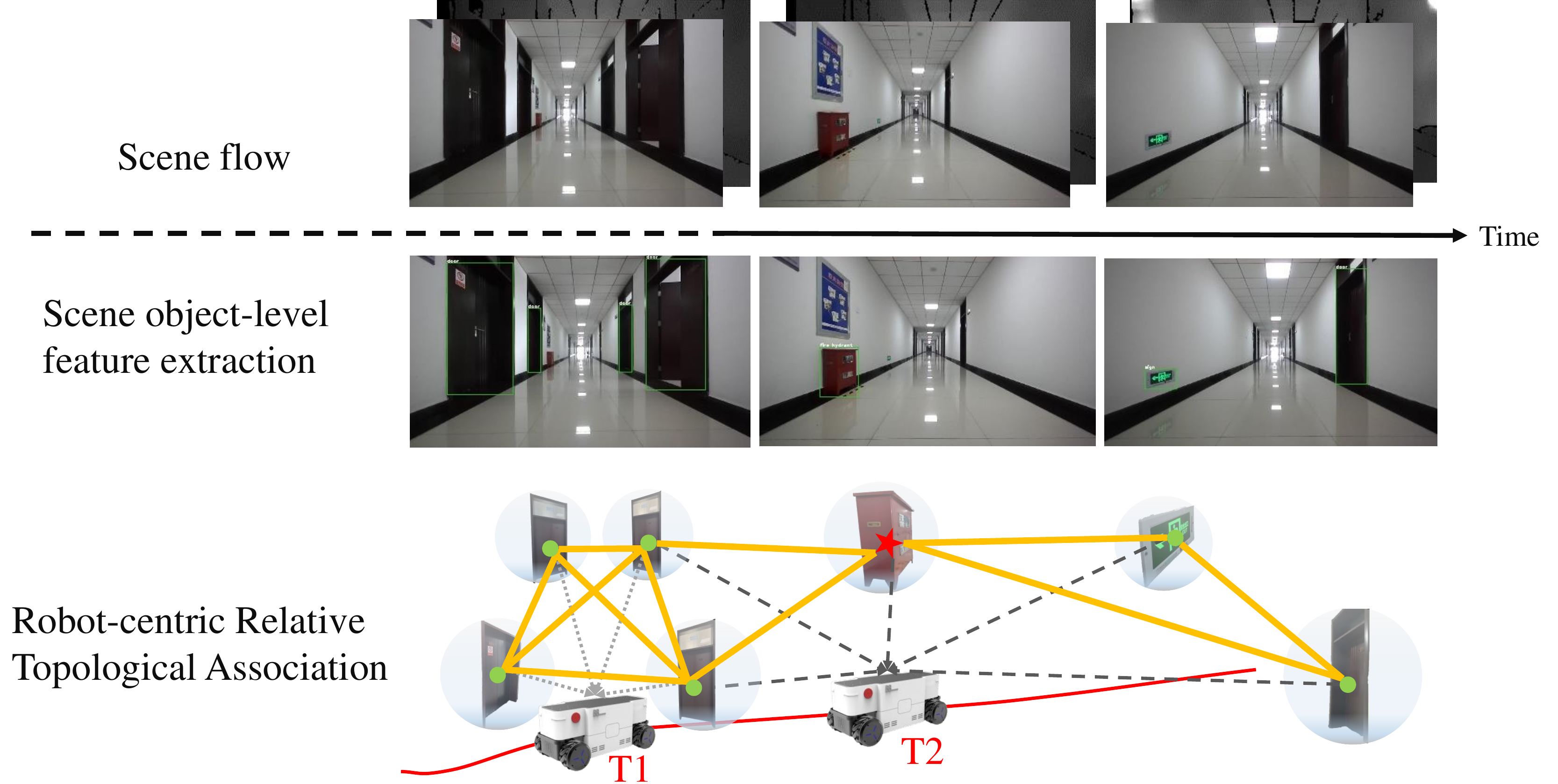}
	\end{center}
	\caption{Schematic representation of the our proposed mapping system in scene flow. }
	\label{fig:1}
\end{figure}
%\end{comment}
With these metric maps, mobile robots obtain precise localization and navigation. 
However, the expensive computation, incremental accumulation errors and vast storage of metric maps challenge its geometrical consistency and make it's more difficult to maintain the accuracy of the map.
%On the one hand, global consistency is not easily guaranteed, due to factors such as require storing a large amount of date as the map-scale extensions and prone to accumulation of bias~\cite{Alpher06}.
%On the other hand, visual appearance-based global localization often suffer from some challenges such as illumination variations and new dynamic objects~\cite{Alpher02}~\cite{Alpher05}. 
Especially, this task can be particularly complex when done online on a robot with limited computing capabilities and storage resources~\cite{c2}. 
%Therefore, it is of great significance to carry out the lightweight mapping and robust localization.
%Vision-based mapping and localization has recently received great attention due to the low cost of cameras and the richness of the sensor data provided.
%Despite the significant progress over the past decades, the robustness of vision-based mapping and localization is still a intractable question, since it has limitations in following conditions, such as dynamic, changing and large scale environment.
%Especially, this task can be particularly complex when done online on a robot with limited computing capabilities and storage resources.
Thus, robust and lightweight mapping and localization is of special interest for both academic and industry research.

%Generally, there are two fundamental types of environment representation: metrical and topological map.
%Classical metric maps build precise geometric model of environment and clearly correspond to the real world, which based on Simultaneous Localization and Mapping(SLAM) technology.
%With these metric maps, mobile robots obtain precise localization and navigation.
%However, the expensive computation, incremental accumulation errors and vast storage of metric map challenge its geometrical consistency and make it's more difficult to build high precise map.
%Thus, typical metric maps unlikely to optimally fit to complex large-scale and long-term real-world navigation.
%Different from metric map, topological method models the environment as a graph of discrete locations, where nodes represent distinctive places in the environment and edges indicate the relationships between them.
%Existing research shows that topological maps are lightweight and more suitable for large-scale mapping and localization.

For the above problems, the topological approach
provides a lightweight mapping solution similar as simple and compact, scale better and require much less space to be stored than metric map~\cite{c3}. 
%Topological maps represent an environment as an abstract graph  with nodes and edges encoding places and adjacencies~\cite{Alpher06}, so that it is not suffer as directly from the accumulation of movement errors. 
Despite of their remarkable results, there still some challenges in topological mapping methods.
For example, pure topological map is not sufficient for robot navigation, which calls for metric information.
Besides, most existing methods build the topological graph based on the global consistent metric map.
They don't avoid the disadvantages of metric map completely.
These above reasons motivate us to design a real lightweight and robust mapping and navigation method.
These two tasks become particularly challenging when the environmental condition changes due to dynamic people, objects, and/or when the scale of the environment becomes very large.

LiTang et al. proposed a topological local-metric framework in which the global coordinates does not exist to achieve long term mapping and localization~\cite{c4}.
However, its integrated visual appearance-based loop closure method reduces robustness to environmental changes.
Recently, semantic graph description provides an effective method for accurate and robust localization.
Existing work~\cite{c5}~\cite{c6}~\cite{c7}~\cite{c19} have explored that high level semantic features provide a more robust representation for the scene since they incorporate the information of objects’ own properties and their mutual relations. 
They are able to cope with the global localization under extreme appearance changes successfully~\cite{c5}. 

In this paper, we combine the above advantages to realize a lightweight mapping that organizes the semantic and geometric properties of environmental objects through topological graph, and completes a robust localization based on semantic graph description of the local scene, we calls it an Object-level Topological Semantic Map(OLTSM).
%Mapping and localization methods based on topological graph and semantic fusion are also being explored.
%YuLiu et al. proposed a framework to integrate dense semantics, 3D topology, graph matching and 3D alignment into a object level global localization algorithm~\cite{Alpher10}, which demonstrates both high accuracy and robustness under drastic appearance variations. 
%However, it needs to build a dense 3D semantic map, so that it is not suitable for long-term navigation.
%Ygor C. N. Sousa et al. proposed a topological semantic mapping by consolidation of deep visual features~\cite{Alpher23}. However, the algorithm requires a lot of training data to improve its generalization ability, and it cannot be applied in real situations.
%Inspired by these excellent topological semantic SLAM systems, 
%Inspired by these excellent mathods, 
%we proposd an Object-level Topological Semantic Map(OLTSM), a lightweight mapping and long-term robust localization framework. 
An overview of the framework is shown in Fig~\ref{fig:1}. 
%OLTSM represents an environment as a lightweight abstract graph, based on 2D object detection technology and depth information, topological nodes are described using their own semantic properties of object-level landmarks, and topological edges are described using the semantic association properties between object-level landmarks. 
OLTSM represents an environment as a lightweight abstract graph, topological nodes are object-level landmarks, and topological edges are described using the semantic association properties between object-level landmarks.
Extract semantic features of objects and relative semantic associations between objects in a visual odometry (VO)-like manner by means of defined learning topology primitives.
Particularly, inspired by the way humans navigate, we build a map with the robot ontology as the coordinate origin, to control the offset error between local neighboring nodes.
%Global consistency of the topological map is achieved using magnetic declination and coordinate invariance of vectors.
Meanwhile, a hierarchical memory management strategy was introduced to improve the efficiency of online mapping.
Furthermore, we introduced an object-level semantic graph descriptor and a semantic graph matching method to achieve robust localization.
Last but not least, the algorithm was deployed to a low-cost embedded platform and tested it in the muilt-session real world. 

The main contributions are summarized as follows:
\begin{itemize}
%\item An online lightweight mapping solution based on semantic association and topological organization is proposed. This map is analytically shown based on the robocentric coordinate.
%\item To adapt to the dynamic changes of the environment, a semantic scene graph descriptor is introduced to realize the object-level qualitative description of the surrounding environment. 
%\item The scene similarity measurement method based on graph matching and the use of effective memory management realize a fast and robust localization method.
%\item Online mapping amd localization based on OLTSS is realized on the low-cost embedded device Jetson Nano with TensorRT deployment.
%\item The effectiveness of the proposed solution is verified by the autonomous operation of the mobile robot in the real world. In particular, Our approach greatly outperforms classical metric map methods in terms of lightness on the collected dataset. And also the proposed method obtains higher robustness than the method based on visual appearance information in challenging situations.
\item An online lightweight mapping solution based on semantic associations of objects and topological organization is proposed. This map is built in a robot-centric human-like navigation way.
%\item Multiple attribute information of objects including semantic, geometric center, and object-level semantic scene graph as descriptors that can be used to achieve accurate object-to-object semantic associations.
%\item The scene similarity measurement method based on graph matching and the use of effective memory management realize a robust localization method that is suitable for long-term dynamic environmental changes.
\item A localization method based on object-level semantic scene graph matching is proposed, which is robust to variations in dynamic environmental such as viewpoint and illumination.
\item Online mapping and localization based on OLTSM is realized on the low-cost embedded platform.
\item Experimental results validate the effectiveness of our methods. Our approach greatly outperforms classical metric mapping methods in terms of lightweight. 
And also the proposed method obtains higher robustness than the method based on visual appearance information in long-term dynamic situations. 

\end{itemize}

%---------------------------------------------
\section{Related Work}

%Mapping and localization is one of the longest running research areas in mobile robotics, as key functions for performing autonomous navigation tasks. 
In this section, we review the related work on mapping and localization methods. 

For mapping, classical metric mapping methods have reached maturity by accurately encoding the geometric 3D information about the environment.
However, these methods still face challenges for long-term navigation in large, unstructured and dynamic environments.
%Such as some methods represent the environment as precise and sparse 3D landmarks by specific determination features (eg points, lines)[ORBSLAM][][].
%However, these sparse landmarks cannot perform navigation tasks. 
%Contrarily to sparse representations, dense representations attempt to provide high-resolution models of the 3D geometry [dense map][][][]. 
%Although these models are more suitable for obstacle avoidance and path planning for mobile robots, they also typically require the storage of large amounts of data []. 
%Visual sensors (eg,stereo, RGB-D) also facilitate the construction of highly accurate geometric maps, such as occupancy grid maps[].
%However, it is not easily maintained and built in the long term navigation.
For above reason, topology maps are widely explored to represent the environment in abstract graphs, achieving a simple and compact lightweight representation while control the error between local adjacent nodes~\cite{c8}~\cite{c9}~\cite{c10}~\cite{c11}. 
%The solution is adequate when only topology-level navigation is required, and is not suitable for robot navigation that requires metric guidance[2019tc].
However, the solution is not suitable for robot navigation that requires metric guidance~\cite{c4}.
Therefore, the one idea is the construction of highly consistent topological representations based on metric mappings~\cite{c9}~\cite{c10}. 
%Navigation relies on precise absolute poses between adjacent nodes provided by the global metric map. 
Although it makes the map lightweight, it also limits the high scalability of topological maps and is not easily applicable to long-term navigation. 
The other idea is incorporating accurate local metric information in topological representations~\cite{c4}~\cite{c11}. 
%Navigation relying on relative poses between adjacent nodes provided by the local metric map. 
However, local matching localization using image level is still not sufficiently robust when the environment changes.
Moreover, due to the widespread successful application of deep learning, the learning-based approach has also attracted a lot of interest~\cite{c12}~\cite{c13}. 
It relies on a large number of labeled datasets and cannot be applied well to an unfamiliar environment.

For localization, the existing methods of localization can be divided into two major categorys, based on visual appearance information or spatial relationships of landmarks.
%A common visual appearance based method is visual feature matching.
In the last decade, a large number of methods based on visual appearance information have been proposed that give reliable performance under perceptually similar conditions~\cite{c14}~\cite{c15}.
%However, when the viewpoint change is large, the localization systems become less reliablel.
%Futhermore, to improve accuracy and robustness, several extensions have been proposed, such as by using convolutional neural networks (CNN) [2021his157]and fusing semantic and appearance information[slam++][2020VIL][cubeslam].
To futher improve accuracy and robustness, several extensions have been proposed.
The one idea is to extract and describe visual features by using convolutional neural networks (CNN)~\cite{c16}.
%In [2021his157], the viewpoint invariant landmarks are generated by CNNs. 
However, when the viewpoint change becomes significant, these visual landmarks also become unreliable.
%Several uses semantics and appearance information methods are proposed to above problem[slam++][2020VIL][cubeslam].
The other idea is to integrate semantic and appearance information~\cite{c17}~\cite{c18}~\cite{c19}.
However, they focused on reducing drift using image-based associations, our approach tend to perform localization through a graph-based semantic local environmant representation model.  
Another localization methods by using the spatial relationship of surrounding landmarks to represent, such as the graph-based methods~\cite{c5}~\cite{c6}~\cite{c7}.
Futhermore, random walk~\cite{c5}~\cite{c21}, graph kernel~\cite{c20}, histogram~\cite{c7} and graph embedding~\cite{c6} algorithms are widely used to extract the information of the graph structure and perform similarity matching.
YuLiu et al. proposed to utilize graph matching and 3D alignment into a object-level global localization algorithm~\cite{c21}. 
It was demonstrated that graph-based object-level semantic information descriptors which can improve the localization performance.

Driven by above methods and inspired by the navigation behaviors of human being, in this letter, we intend to incorporate the semantic and geometric information into the graph structure. 
Then, the similarity calculation between semantic scene graphs is realized by graph matching.

%--------------------------------------------------------------
\begin{comment}
\section{SYSTEM OVERVIEW}

We propose a lightweight and robust map construction and localization method for anthropomorphic path search and long term navigation. 
This map is based on object-level landmarks as navigation nodes, and organizes object-level landmarks and semantic associations between landmarks based on a graph topology. 
Particularly, we construct an efficient topological semantic mapping and localization framework, including object detection, topological semantic mapping, and semantic scene graph description and graph matching, which can be executed in real-time with TensorRT deployment on the embedded device Jetson Nano.

Our framework is composed of two main modules: map construction and localization. 
In map construction module, we mainly accomplish the extraction of semantic information, as well as realize the extraction of environmental topology information centered on the robot ontology through the representation and association of object-level landmarks. 
In localization module, on the one hand, we accomplish an object-level semantic scene graph descriptor, which used random walk method to descriptor local topology information of objects. 
On the other hand, the scene similarity measurement method based on graph matching and the use of effective memory management realize a robust localization method taht is suitable for long-term dynamic environmental changes. 
Fig 1 outlines the specific modules and the connections between them.
\end{comment}

\begin{figure*}[htbp]
	\begin{center}
		\includegraphics[width=1 \linewidth]{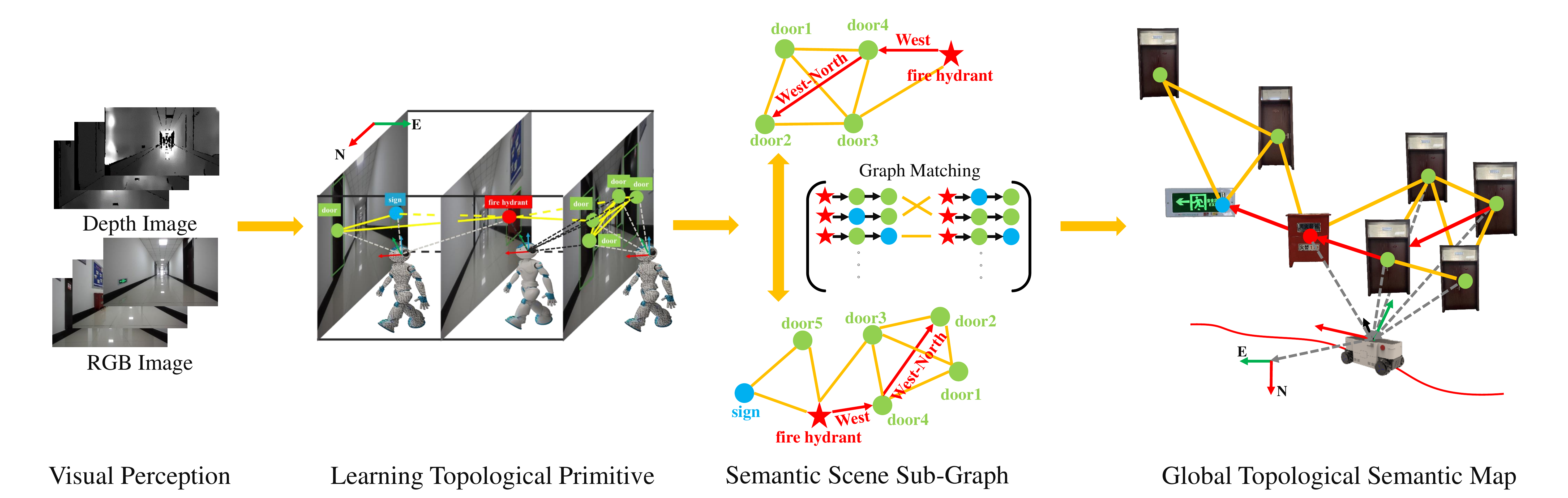}
	\end{center}
	\caption{Overview of the our proposed mapping system. Inspired by the way humans navigate, our method takes setereo camera as visual perception, and then implemented semantic feature extraction and robot-centric topology-based object-level semantic association by the proposed visual odometry-like process of learning topological primitives. On this basis, a hierarchical memory management mechanism improves the efficiency of online mapping with limited computational resources. In the process of localization, semantic scene graphs are used to augment the described objects. Finally, localization is achieved by semantic scene graph matching.}
	\label{fig:2}
	%\label{fig:onecol}
\end{figure*}
%-------------------------------------------------------------------------
\section{Lightweight Object-level Topological Semantic Map}

% The map is a fundamental representation of aspects of interest (e.g., landmarks, obstacles) describing the environment in which the robot operates~\cite{Alpher10}. 
The map is a fundamental representation of interest (e.g., landmarks, obstacles) describing the environment in which the robot operates~\cite{c21}. 
Inspired by the way humans navigate, we pay more attention to recent relative movements than to the global position  for long-term navigation.
%Thus, the environment is represented as a lightweight abstract topological graph in this paper. 
Therefore, in this paper, we represent the environment by using a lightweight abstract topology graph that records the relative associations between objects.
The basic structure of this map is a graph defined as \begin{math}
G = \left\lbrace N,E\right\rbrace 
\end{math}, where \begin{math}
N
\end{math} and \begin{math}
E
\end{math} denote the nodes and edges of the graph, respectively.
%On this basis, a  without global poses map with the robot ontology as the origin of the coordinates is constructed by using topological organization strategy.
%On this basis, a without global poses map with the robot-centric is constructed by using topological organization strategy. 
On this basis, A robot-centric global-free pose map is constructed using topological organization.
%In particular, multiple attribute information of objects is extracted to achieve an effective node description. 
In particular, semantic features of objects and data associations between objects are extracted though a VO-like process, which we define as learning topological primitives.
The details are described as follows.
%Furthermore, to control the accumulated offset error, we propose a map without global poses and with the robot ontology as the origin of the coordinates.

%------------------------------------------------------------------------
\subsection{Topological Map Representation}

%In this section, we present a concrete representation of the constructed abstract graph, as follows.
In this section, we first introduce the concrete representation of the constructed abstract graph as follows. The representation of the map is shown in the semantic scene sub-graph in Fig~\ref{fig:2}.
\subsubsection{Node Representation}

%For topological nodes, the object's own semantic features are used to define, such as class, color, etc. 
In this paper, we take the objects in the scene as topological nodes and define the nodes with the semantic properties of the objects themselves, such as class, color, etc.
Thus, for each node \begin{math}N_{i}\end{math} belonging to $ N $, the corresponding properties are defined as \begin{math}N_{i}=\left\lbrace ID,class,center,n_{i}\right\rbrace \end{math}. 
\begin{math}ID\end{math} is the serial number added to the graph in order. \begin{math}center\left(x,y,z\right) \end{math} is the coordinate of the object center point obtained by fusing deep imfromation. 
\begin{math}n_{i}\end{math} is defined as a proxy for additional properties of the node. 
For example, functional and operational properties, and 6D pose (position and orientation), etc.
Peculiarly, to achieve an exact match between them, different with hand designed descriptor, we propose a novel multiple attribute semantic descriptor for object-level nodes including semantic, geometric center, and object-level semantic scene graph with topological organization.
%The semantic information and semantic scene graph information are obtained as described in Section \Rmnum{4}.B and Section \Rmnum{4}.C, respectively.
The semantic scene graph descriptors are obtained as described in Section \Rmnum{4}.A.

%------------------------------------------------------------------------
\subsubsection{Edge Representation}

For topological edges, the associated object-level node relative relationship attributes are used to define, for example, the relative direction and distance between nodes. 
Thus, for each edge \begin{math}E_{ij}\end{math}, connecting the neighboring nodes \begin{math}N_{i}\end{math} and \begin{math}N_{j}\end{math}, belonging to \begin{math}E\end{math}, the corresponding properties are defined as \begin{math}E_{ij}=\left\lbrace dis_{ij},yaw_{ij},e_{ij}\right\rbrace \end{math}. \begin{math}dis_{ij}\end{math} is a rigid relative distance between node \begin{math}N_{i}\end{math} and \begin{math}N_{j}\end{math}, which is also used as the weight of the topology graph. 
It can be obtained in creating the map from a variety of sources, such as sensor measurements and human measurements.
\begin{math}yaw_{ij}\end{math} is a relative direction between node \begin{math}N_{i}\end{math} and \begin{math}N_{j}\end{math} in the geomagnetic coordinate system. 
we introduce IMU to calculate the magnetic declination angle.
%, and make the declination error less than 2 degrees by optimize. 
\begin{math}e_{ij}\end{math} is defined as a proxy for additional properties of the edge. 

%------------------------------------------------------------------------
\subsection{Robot-centric Relative Topological Association}

%In this paper, we present the learning topological primitive for semantic association between objects.
%In this section, we first present the learning topological primitive for constructing object-level abstract topological graphs.
%The definition of a learning topological primitive is given, which implements learning-based scene object-level feature extraction $ V_{f}^{obj} $ and topology-based data association $ \bigoplus $ through a VO-like process in a stream of scenes during a continuous time period. 
The purpose of this section is to perform scene object feature extraction and relative topological association for constructing object-level abstract topology graphs.
For this purpose, we propose a learning topological primitive $ LTP $, which is defined to implement learning-based object feature extraction with fusion of spatial and semantic information $ V_{f}^{obj} $ and topology-based data association $ \bigoplus $ through a VO-like process in a stream of scenes during a continuous time period $ T $.  
The formula is represented as follows:
\begin{equation}
\begin{aligned}
%LTP=Sg^{T1}\left ( V_{1f}^{obj1},V_{1f}^{obj2},\cdots, V_{1f}^{obji} \right )\bigoplus \\
%Sg^{T2}\left ( V_{2f}^{obj1},V_{2f}^{obj2},\cdots, V_{2f}^{objj} \right)
LTP=\left ( V_{f}^{obj1} \bigoplus V_{f}^{obj2}\bigoplus\cdots\bigoplus V_{f}^{obji} \right )^{T}
\end{aligned}
\end{equation}
Then, we refine the relative association between objects by the principle of vector coordinate invariance. 
The left half of Fig~\ref{fig:2} shows the VO-like process from visual environment perception to learning topological primitive construction.
%In this paper, we extract information about multiple attributes of objects as descriptors to achieve an exact match between them. 
%A key task for topological representation of the environment is effective semantic detection of object-level nodes. 
%-------------------------------------------------------------------
\subsubsection{Learning Topological Primitive}
%With the robot moves in scene flow, the first task is effective semantic feature extraction of object, thus, we based on the lightweight 2D object detection method YOLOv5[] that achieves advanced performance to extract semantic information about the environment's objects. 
With the robot moves in scene flow, in this letter, we first extract semantic information about the environment's objects based on the lightweight 2D object detection method YOLOv5\footnote{https://github.com/ultralytics/yolov5} that achieves advanced performance.
We choose long-term static objects that are more aligned with human navigation road signs as valid landmarks, such as door, fire hydrant, pillar etc.
In addition, since the 3D center point of an object is subject to less affected variation in viewpoint, we implemented 3D center point detection based on the object 2D detection box by adding depth information.
%However, the detection accuracy of YOLOv5 in the real environment is still limited, so some data for training based on the data format of COCO[] dataset are added.

Afterwards, the relative positions between objects in the scene flow are associated in a robot-centric topology.
The association of environmental topology information is shown in Fig~\ref{fig:3}.a.
%In the constructed topological map, we use the robot ontology as the coordinate origin, and by transforming the nodes and the robot ontology coordinates, as well as the robot ontology coordinates and geomagnetic coordinates, the relative direction between adjacent nodes is expressed as the direction vector  \begin{math}\vec{d}=\left(x_{2}-x_{1},y_{2}-y_{1}\right) \end{math} in the calculation. 
In the constructed topological map, we use the robot ontology as the coordinate origin, and by transforming the nodes and the robot ontology coordinates, the relative direction between adjacent nodes is expressed as the direction vector  \begin{math}\vec{d}=\left(x_{2}-x_{1},y_{2}-y_{1},z_{2}-z_{1}\right) \end{math} in the calculation. 
Therefore, any node in the graph and its neighbors maintain only a relative relationship, which reduces the impact of global errors.
%Fig 3 visualizes the contrast of detection performance in environments where visual feature-based methods and semantic object-based methods differ greatly in perspective, appearance, and layout, respectively.

%------------------------------------------------------------------------
\subsubsection{Refinement of Relative Topological Association}

\begin{figure}[htbp]
	
	\begin{minipage}{0.48\linewidth}		
		\centerline{\includegraphics[width=4.3cm]{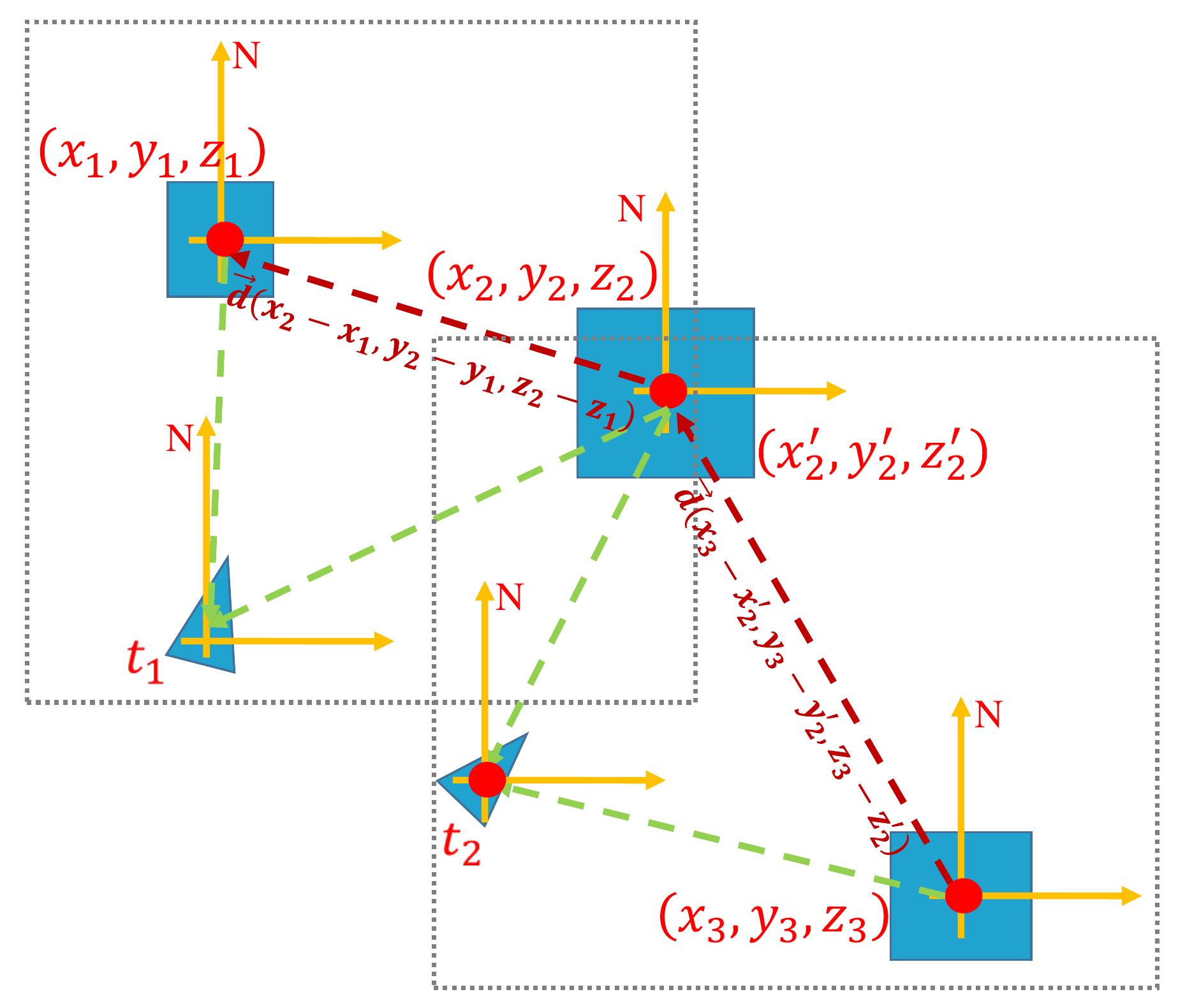}}		
		\centerline{(a)}		
	\end{minipage}	
	\hfill	
	\begin{minipage}{.48\linewidth}		
		\centerline{\includegraphics[width=4.3cm]{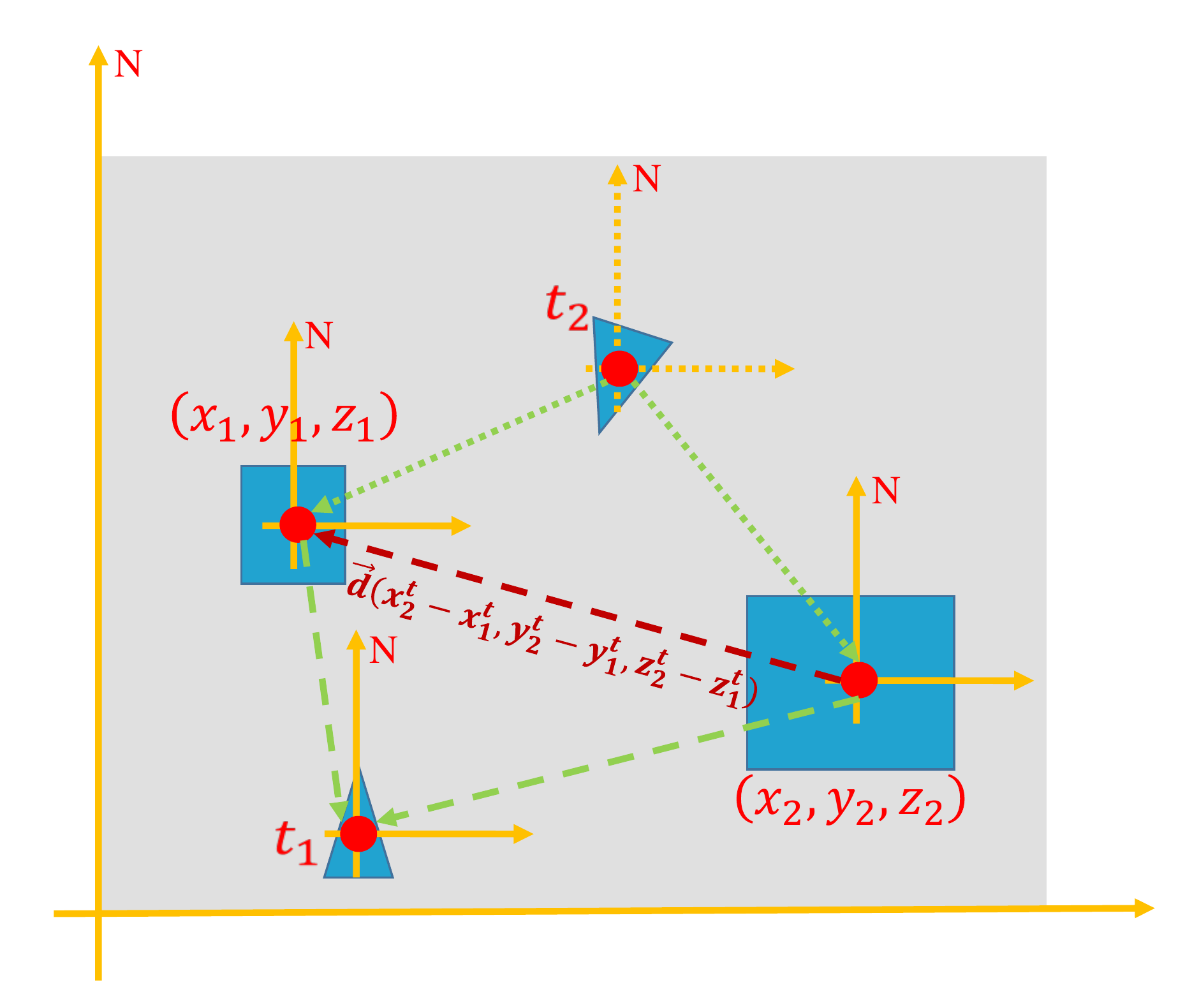}}		
		\centerline{(b)}		
	\end{minipage}
	\caption{Illustrates of the robot-centric relative topological association.}
	\label{fig:3}
	%\label{fig:onecol}
\end{figure}

%To ensure effective localization and navigation through the constructed topological semantic map without global poses, we propose a strategy for organizing global topological information.
%To ensure effective localization and navigation through the constructed topological semantic map without global poses, we propose a strategy to refine the relative positions between objects by transforming the robot ontology coordinates to geomagnetic coordinates based on the principle of vector coordinate invariance.
To ensure effective localization and navigation through the constructed topological semantic map without global poses, we propose a strategy based on the principle of vector coordinate invariance to refine the relative positions between objects by transforming the robot body coordinates to geomagnetic coordinates.
Since the direction of the geomagnetic coordinate system is usually constant, based on the principle of coordinate invariance of vectors, the relationship between adjacent fixed nodes does not change with time and space, as is shown in Fig~\ref{fig:3}.b. 
%This robot-centric topological organization strategy establishes the basis for long-term localization and navigation based on this map. 
The specific implementation of the pseudocode is shown in Algorithm 1.
\makeatletter  
\def\BState{\State\hskip-\ALG@thistlm}  
\makeatother 
\begin{algorithm}[htbp]  
	\caption{Robot-centric Topological Association}  
	\label{alg:Framwork}  
	\begin{algorithmic}[1]  
		\Require  
		Adjacent landmarks $ N_{1} $ and $ N_{2} $, given the coordinates $ CN_{12} $ of $ N_{1} $ in the camera coordinate system at the moment $ t_{2} $, solve the coordinates $ BN_{22} $ of $ N_{2} $ in the robot coordinate system at the moment $ t_{2} $;      
		\State $ CN_{11}=\left ( x_{1},y_{1},z_{1}\right ),CN_{21}=\left ( x_{2},y_{2},z_{2}\right )\leftarrow $  The coor-dinates in the camera coordinate system at the moment $ t_{1} $;  
		\label{code:fram:extract}   
		\State $ \left [ R,T\right ] $ Conversion matrix of camera coordinate system to robot rigid body coordinate system;  
		\State $ BN_{11}=R\ast CN_{11}+T,BN_{21}=R\ast CN_{21}+T,$ $BN_{12}=R\ast CN_{12}+T $; 
		\BState \emph{\textbf{Refinement}}:
		$ yaw_{1},yaw_{2}\leftarrow  $ The deflection angles of the robot's rigid body and magnet moments $ t_{1} $ and $ t_{2} $, respectively;
		\State $ MN_{11},MN_{12},MN_{21} 
		\leftarrow \quad \qquad\quad \qquad\quad \qquad$
		$ MN_{it}.x=BN_{it}.x\ast \cos \frac{yaw_{t}\ast \pi  }{180} + BN_{it}.y\ast \sin \frac{yaw_{t}\ast \pi  }{180} $;
		$ MN_{it}.y=BN_{it}.y\ast \cos \frac{yaw_{t}\ast \pi  }{180} - BN_{it}.x\ast \sin \frac{yaw_{t}\ast \pi  }{180} $;
		\State $ \vec{d}=\left ( MN_{21}.x_{2}-MN_{11}.x_{1},MN_{21}.y_{2}-MN_{11}.y_{1}\right ) $;
		\State $ \mathbf{MN_{22}}=\mathbf{MN_{12}}+\vec{d} $;
		\State $ BN_{22}\leftarrow \mathbf{MN_{22}},yaw_{2} $;
	\end{algorithmic}  
\end{algorithm}

%-----------------------------------------------------------------------

\subsection{Hierarchical Memory Management}

%To improve the efficiency of online map updates and localization with limited computing resources, we have introduced a memory management mechanism [].
%Before starting to perform localization, 
%Finally, to improve the efficiency of online mapping with limited computing resources, in this section, we have introduced a bottom-up hierarchical memory management mechanism.

\begin{figure}[]
	\begin{center}
		\includegraphics[width=1\linewidth]{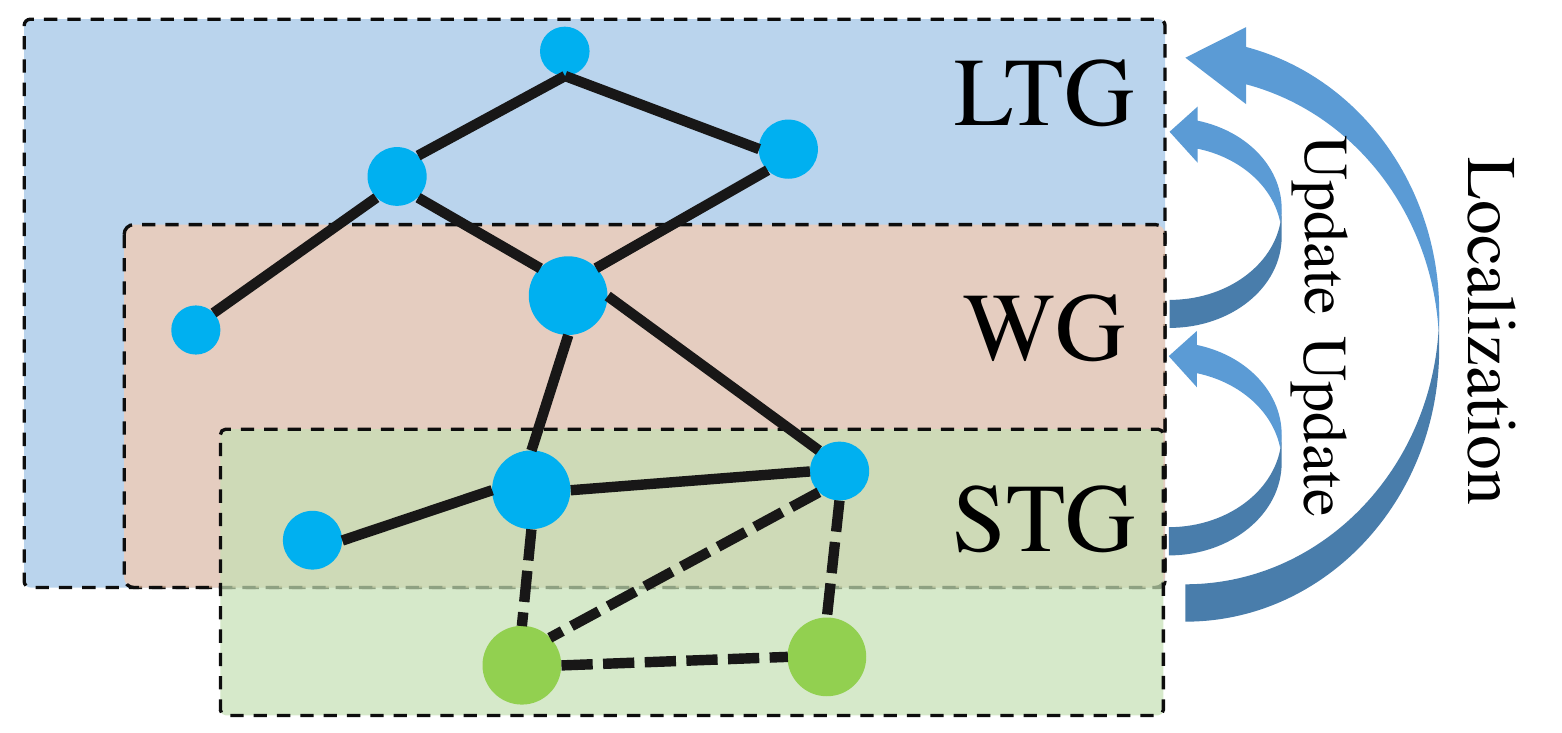}
	\end{center}
	\caption{Schematic representation of the hierarchical memory management. Blue dots are nodes already in the graph, green dots are newly associated nodes. Update the graph with a bottom-up hierarchy.}
	\label{fig:4}
	%\label{fig:onecol}
\end{figure}

Finally, to improve the efficiency of online mapping with limited computing resources, in this section, we have introduced a hierarchical memory management mechanism, which reduces the retrieved area through a bottom-up hierarchical matching update strategy.
Thus, the memory management is divided into three hierarchicals in this paper: Short-Term Graph (STG), Working Graph (WG) and Long-Term Graph (LTG), as shown by Fig~\ref{fig:4}.
%STG is a temporary local sub-graph constructed from previously added nodes and objects in the current frame, it has a fixed size 5.  
%The WG is a local sub-graph with nodes in STG as vertices and a random walk with three steps as the radius. 
%LTG is a global map that has been constructed.
%For map update, by matching STG and WG, STG nodes that are not in the WG will be added to the LTG, and new spatial semantic association information is constructed.  
%For localization, by matching STG and LTG, we can fast determine the current location of the robot in the LTG. 
STG is a bottom temporary local sub-graph constructed from previously added nodes and objects in the current frame sequence, being fixed to 5 nodes.
WG is a middle-level local subgraph constructed by randomly walking around with a radius of three steps with the node in STG as the root node.
LTG is a top-level global graph that has been constructed.
To satisfy online map updata, by leveraging the graph matching method proposed in Section \Rmnum{4} to quickly match the local semantic scene subgraphs STG and WG, STG nodes that are not in the WG (nodes in STG but not in WG) will be added to the LTG, and new spatial semantic association information is constructed. 
%In other words, this reduces the search area for matches.
%In addtion, when a loop-closure is detected or re-localization is required, by matching STG and LTG, we can also fast correct the map and determine the current location of the robot in the LTG.
In addition, when a re-localization is required, by matching STG and LTG, we can also determine the current location of the robot in the LTG.
%\begin{comment}
\begin{figure}[]
\begin{center}
\includegraphics[width=1\linewidth]{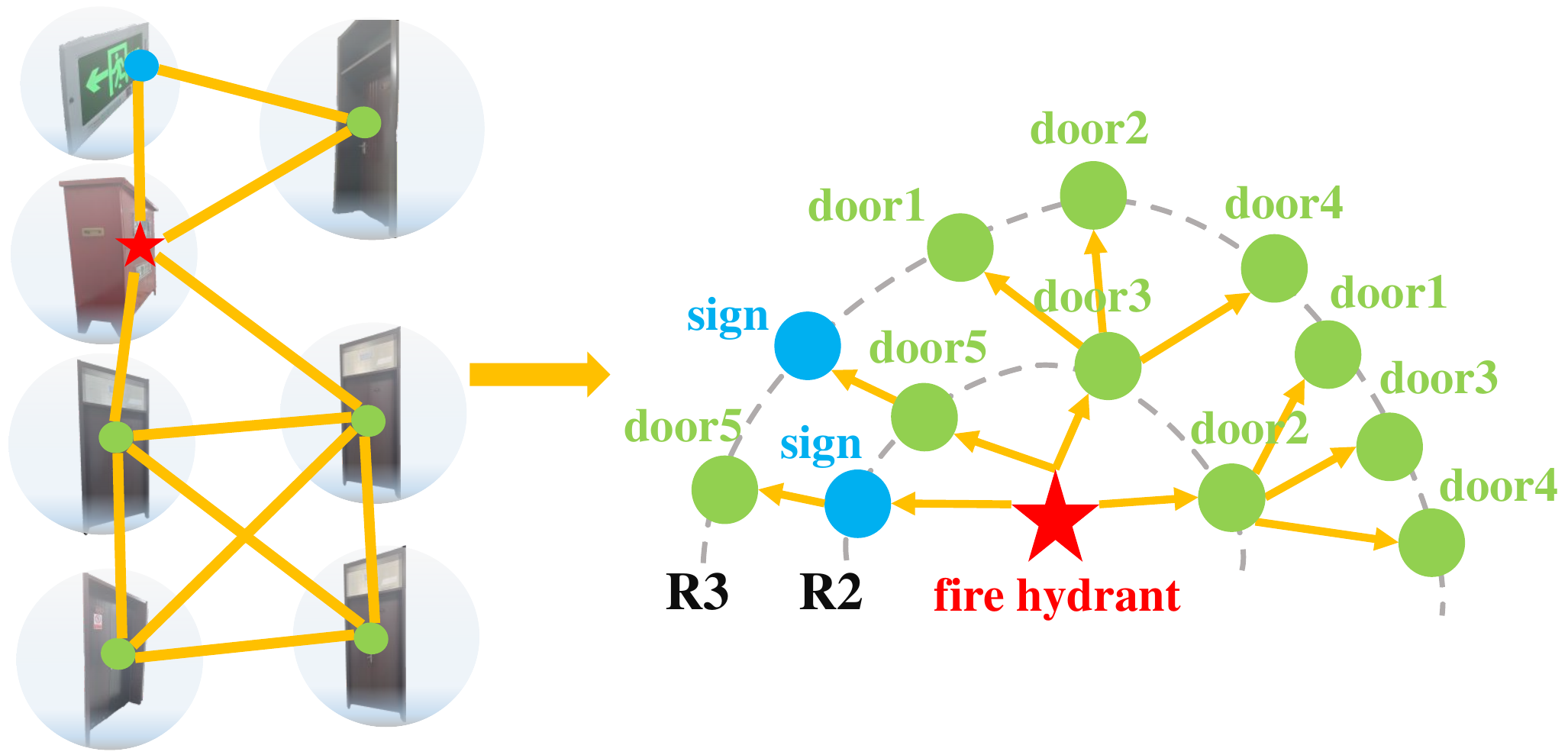}
\end{center}
\caption{Schematic representation of the semantic scene graph descriptors.}
\label{fig:5}
%\label{fig:onecol}
\end{figure}
%\end{comment}

%------------------------------------------------------------------------
\section{Localization with object-level semantic scene graph matching}

% Localization are the key steps to perform robust long-term navigation task.
Localization is widely acknowledged as crucial for robot navigation, especially for long-term navigation.
Different from previous geometric solutions, that are less robust to environmental changes, 
in this section, we present an online topology localization method based on the above constructed map, which is robust to long-term dynamic environmental changes.
%in this section, we present an online topology localization method based on the above constructed map, by xxxxx, which is robust to long-term dynamic environmental changes.
% In this section, based on the  above topology graph constructed by learning topological primitives, we propose an online robust localization method based on graph matching that is applicable to long-term dynamic environmental changes.
The details are described as follows.
%----------------------------------------------------------------------------
\subsection{Semantic Scene Graph Descriptors}

To improve the accuracy and robustness of localization, we have enhanced the description of each node in the graph. 
Inspired by the~\cite{c6} graph descriptor, we introduce the random walk descriptor and add semantics and the relative direction information between nodes extracted by learning topological primitives, which implements the object-level semantic scene graph descriptor for the nodes.
Specifically, we select the node being described as the root node, and within a certain detection radius, detect and embed all neighboring nodes and the semantic associations between neighboring nodes in a vector group $Vec(DesS,DesD) $ by random walk.  
The element $ DesS $ in the vector represents the object semantic descriptor mentioned above, which consists of the class, color, etc.
The element $ DesD $ in the vector represents the object direction descriptor, which consists of the relative direction and distance between objects and random walk descriptor of the object. 
The above steps are performed cyclically by replacing the root node until the graph being described is fully explored.
%For each described node, we random walker explore it all connected neighboring nodes and recorded semantic information between adjacent nodes. 
%In this work, for mapping, we use 3 steps as the exploration radius.
%For localization, we use 5 steps as the exploration radius.  
In this paper, the exploration radius for mapping is set as 3 steps, while for localization is 5 steps.
%Therefore, the local topology information of objects is stored in descriptors.
The pseudocode of the three-step semantic scene graph descriptor is shown in Algorithm 2. Fig~\ref{fig:5} illustrates the process with an example.
\begin{comment}
\begin{figure}[]
\begin{center}
\includegraphics[width=1\linewidth]{4.eps}
\end{center}
\caption{Four-wheeled mobile robot platform.}
\label{fig:6}
%\label{fig:onecol}
\end{figure}
\end{comment}
\begin{algorithm}[htbp]  
	\caption{ Semantic Scene Graph Descriptor Extraction.}  
	\label{alg:Framwork}  
	\begin{algorithmic}[1]  
		\Require  
		Given node $ N_{i} $ in semantic scene graph $ G $;    
		\Ensure  
		The descriptors for $ N_{i} $;  
		\State Initialize the semantic descriptor vector $ DesS $;\\
		Initialize the direction vector descriptor vector $ DesD $;  
		\label{code:fram:extract}  
		\For{$ j.th $ in neighbor nodes of $ N_{i} $}  
		\State Record $ N_{i} $ semantic properties $ Li $ and the semantic property $ Lj $ of the $ j.th $ neighbor node of $ N_{i} $;  
		\State Record the direction vector $ Vecij $ between the node’s and $ N_{i} $ ;  
		\State Add $ Vecij $ into $ DesDi $ ;
		\For{$ k.th $ in neighbor nodes of $ j $}  
		\State Record the node’s semantic properties $ Lk $;  
		\State Record the direction vector $ Vecjk $ between the node’s and $ k.th $ ;  
		\State Add $ Vecjk $ into $ DesDi $  ;
		\EndFor
		\State Add $ (Li-Lj-Lk) $ into $ DesS $; 
		\State Add $ DesDi $ into $ DesD $; 
		\EndFor  
	\end{algorithmic}  
\end{algorithm}

%-------------------------------------------------- 

\subsection{Semantic Scene Graph Matching}

In this section, once semantic scene graph descriptors are built for the node, we search associations between query graph and database graph by computing a similarity score between the corresponding graph descriptors to achieve accuracy and robust localization.
In this letter, we propose a multi-constraint semantic scene graph similarity matching method, which consists of three parts: Euclidean distance constraint, Confidence constraint and Direction vector constraint.

The euclidean distance constraint of the corresponding node is the first step in the similarity metric, which aims to eliminate the same node in adjacent frames.
The euclidean distance $ S^{D} $  is calculated as follows: 
\begin{equation}
%$$
S^{D}=\sqrt{\left ( x_{1}-x_{2}\right )^{2}+\left ( y_{1}-y_{2}\right )^{2}+\left ( z_{1}-z_{2}\right )^{2}} %\eqno{(2)}
%$$
\end{equation}

Secondly, confidence constraint is mainly to ensure that candidate sequences have consistent class attributes. The confidence constraint for a single random walk  path $ S^{C} $  is calculated as follows: 
\begin{equation}
%$$
S^{C}=c_{1}k^{R}+c_{2}k^{R-1}+\cdots +c_{R} %\eqno{(3)}
%$$
\end{equation}
$ R $ denotes the random walk search radius, $ k $ denotes the number of overall identifiable object classes, and $ c_{*} $ denotes the object class index number at the $ R $th level.

Lastly, the direction vector constraint is mainly by matching the direction of edges between corresponding sequence nodes and by taking the normalized dot product between two descriptors to determine the unique solution.
The direction vector cosine $ S^{\theta } $ of the corresponding edge and the normalized dot product between two descriptors $ S\left (\textbf{M},\textbf{N} \right ) $ are calculated as follows: 
\begin{equation}
%$$
S^{\theta }= \frac{x_{1}x_{2}+y_{1}y_{2}+z_{1}z_{2}}{\sqrt{x_{1}^{2}+y_{1}^{2}+z_{1}^{2}}*\sqrt{x_{2}^{2}+y_{2}^{2}+z_{2}^{2}}} %\eqno{(4)}
%$$
\end{equation}
\begin{equation}
%$$
S\left (\textbf{M},\textbf{N} \right ) = \frac{\sum_{i=1}^{j} \left (\textbf{M}\textbf{N}^{T} \right ) }{\sqrt{\sum_{i=1}^{j} \textbf{M}^{2}}\times \sqrt{\sum_{i=1}^{j} \textbf{N}^{2}}} %\eqno{(5)}
%$$
\end{equation}
\begin{equation}
%$$ 
\textbf{M}=\left [\textbf{V}_{1i}^{1},\textbf{V}_{1i}^{2},\cdot \cdot \cdot ,\textbf{V}_{1i}^{k} \right ];
%$$
%$$
\textbf{N}=\left [\textbf{V}_{2i}^{1},\textbf{V}_{2i}^{2},\cdot \cdot \cdot ,\textbf{V}_{2i}^{k} \right ]
%  \eqno{(6)}
%$$ 
\end{equation} 
 
$ \textbf{V}_{1i}(x_{1},y_{1},z_{1}) $, $ \textbf{V}_{2i}(x_{2},y_{2},z_{2}) $  are the vectors of the corresponding edges in the NED coordinate system with the body as the origin.
$ \textbf{M} $,$ \textbf{N} $ are the vector groups of these two descriptors respectively.
To ensure that the corresponding edges are facing the same way, therefore the direction $ S^{\theta } $ is greater than or equal to zero.

%---------

\section{Experiment}

In this section, we conduct extensive experiments to evaluate the proposed method well.
Towards this arm, we first collect 5 challenging real-world indoor datasets.
They are captured in a weakly textured indoor long hallway and a large scale complex hospital, as described in section \Rmnum{5}.A.
% Towards this aim, we first explain the implementation details, evaluation datasets and metrics.
% We then compare our xxx localization systems with existing models under both xxx.
% In this section, to evaluate the proposed method well, we perform extensive experiments on several self-made real-world datasets, which include weakly textured indoor long hallways and large scale complex hospital scenes, as detail described in section \Rmnum{5}.A.
We then evaluate our mapping performance in terms of lightweight and accuracy by comparing with traditional occupied grid map~\cite{c23} and state-of-the-art sparse point cloud map (ORB-SLAM3~\cite{c22}).
Finally, we compare our semantic scene graph matching localization method with several appearance-based methods, including traditional and learning methods, to demonstrate the advantages of our method.
All experiments are all performed on a consumer-grade NVIDIA Nano ARMv8 Tegra X1 and 4G memory.

%\begin{comment}
\begin{figure}[]
	\begin{center}
		\includegraphics[width=1\linewidth]{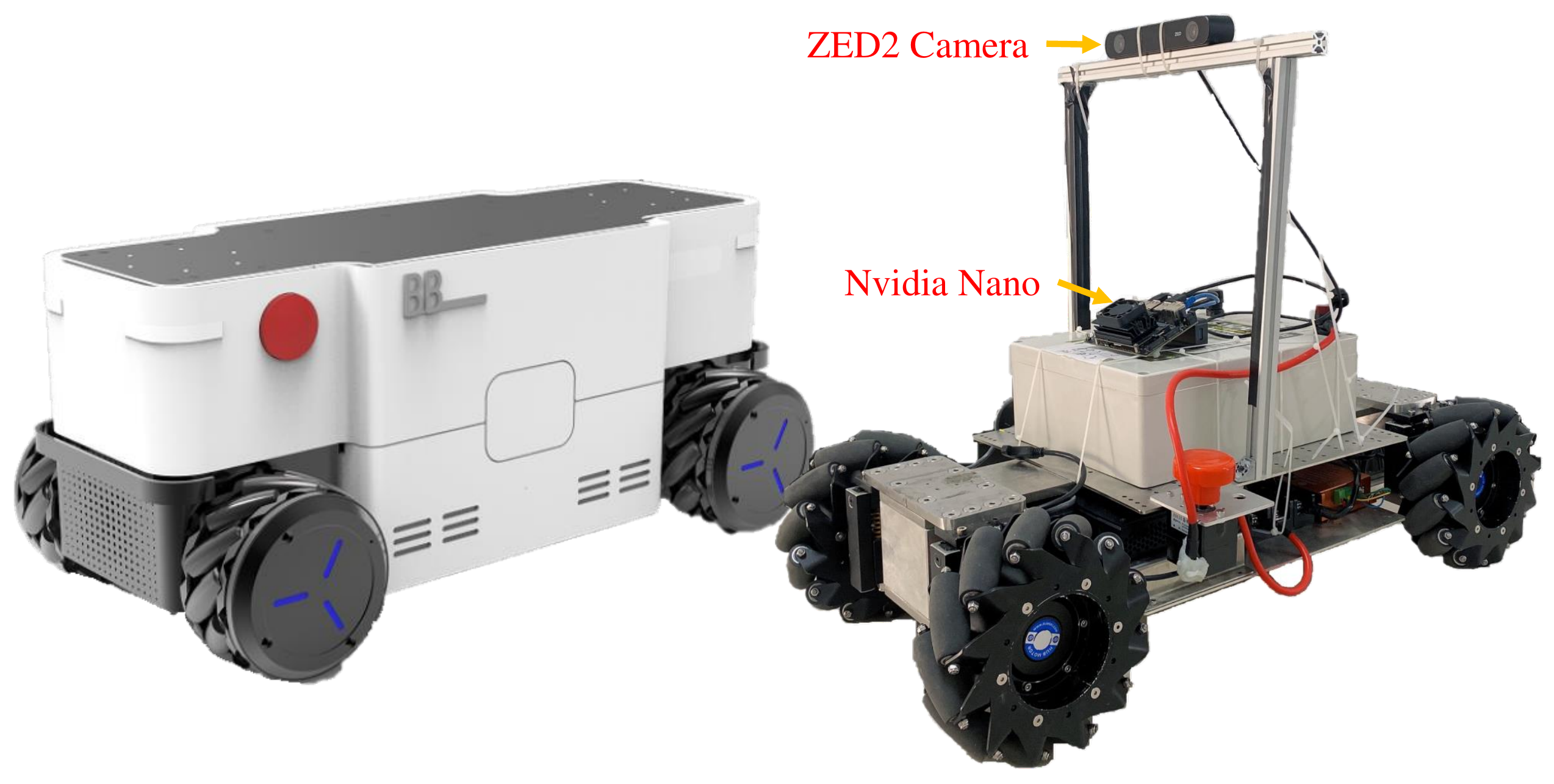}
	\end{center}
	\caption{Four-wheeled mobile robot platform. It is only equipped with an ZED2 stereo camera. All algorithms run on NVIDIA Nano ARMv8 Tegra X1 and 4G RAM.}
	\label{fig:7}
	%\label{fig:onecol}
\end{figure}
%\end{comment}

%----------------------------------------------
\subsection{Benchmark Dataset}

To our knowledge, there does not exist a dataset that meets the requirement of object-level association of our method.

Thus, we create large-scale challenging datasets to .
We collected data by using a four-wheeled mobile robot platform equipped with a ZED2\footnote{https://www.stereolabs.com/zed-2/} stereo camera, as shown in Fig~\ref{fig:7}.

\begin{figure}[htbp]
\begin{center}
\includegraphics[width=1\linewidth]{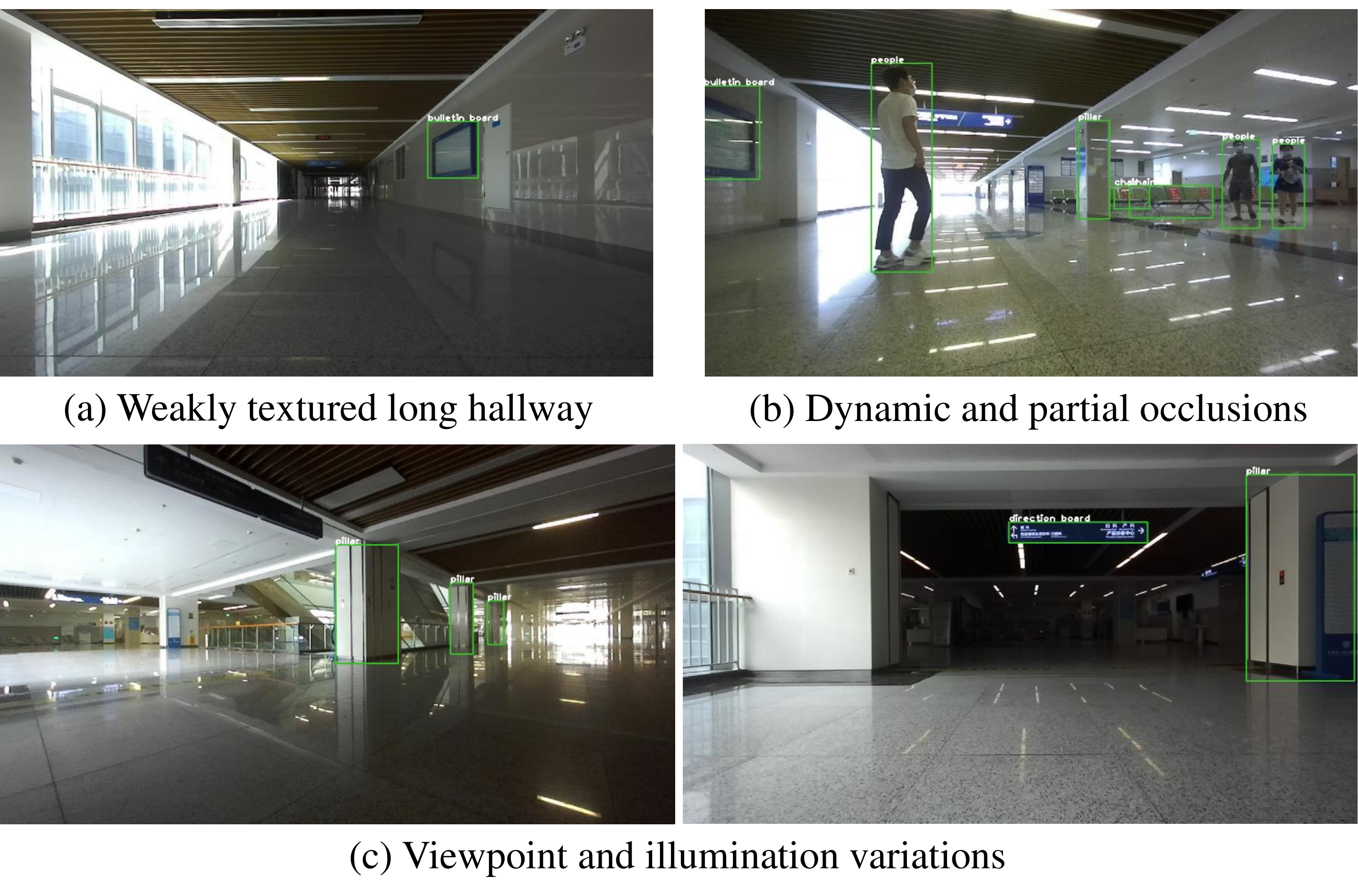}
\end{center}
\caption{The Complex large scale hospital scenes, which include weakly textured long hallways, dynamic and partial occlusion as well as viewpoint and illumination variations, etc.}
%\label{fig:long}
\label{fig:8}
\end{figure}

To meet the testing requirements for algorithm, in weakly textured indoor long hallway with 70 meters length, the test dataset was set to four typical scenarios with significant spatial and temporal variability, including multiple time periods, illumination variations, dynamic and partial occlusions, and large viewpoint variations. 
Details of the long hallway datasets are shown in Table \Rmnum{1}.
In complex hospital scenes, over one thousand square meters of scene data were collected.
Multiple challenging environmental changes are already available simultaneously in the hospital's large scenario dataset, as shown in Fig~\ref{fig:8}.

A total of ten categories of semantic objects are annotated in the two scenes, including doors, fire hydrants, signs, billboards, pillars, chairs, tables, pedestrians, and other objects. 
Fig~\ref{fig:11} and Fig~\ref{fig:8} show the semantic objects detected in the long hallway and hospital scenarios, respectively.
The dataset is publicly available\footnote{https://rec.ustc.edu.cn/share/7a1b2700-2bc7-11ec-ae94-01100e6fec34 password:b52t}.
\begin{table}[h]
	\caption{Overview of the long hallway dataset.}
	\label{table_example}
	\begin{center}
		\begin{threeparttable} 
	    \begin{tabular}{ccccccc}
	    	\hline
	    	& \multicolumn{1}{c|}{}     & \multicolumn{5}{c}{Challenges\tnote{*}} \\ \cline{3-7} 
	    	Sequences & \multicolumn{1}{c|}{Data} & Tl   & Dt   & Lv  & Iv  & Pd  \\ \hline
	    	LH0\tnote{**}      & 21-09-28-10-59            & $\surd$  &      &     &     &     \\ \hline
	    	LH1      & 21-09-28-11-04            & $\surd$  & $\surd$     & $\surd$    &     &     \\ \hline
	    	LH2      & 21-09-28-20-16            & $\surd$  & $\surd$     &     &     &     \\ \hline
	    	LH3      & 21-10-06-11-11            & $\surd$  & $\surd$     &     & $\surd$    &     \\ \hline
	    	LH4      & 21-10-11-09-34            & $\surd$  & $\surd$     &     &     & $\surd$    \\ \hline
	    \end{tabular}
        \begin{tablenotes}
        	\footnotesize               
        	\item[*] Tl, Dt, Lv, Iv and Pd are the abbreviation of texture-less, different time period, large viewpoint variation, illumination variationand and pedestrian and partial occlusion, respectively.  
        	\item[**] LH0 is the baseline dataset and the rest of the sequences are the test dataset.
        \end{tablenotes} 
    \end{threeparttable} 
    \end{center}
\end{table}
%\begin{comment}
%\end{comment}
\begin{comment}
\begin{table}[h]
	\caption{Overview of the long hallway dataset}
	\label{table_example}
	\begin{center}
		\begin{tabular}{ccc}
			\hline
			ID  & Date             & Environments                   \\ \hline
			LH1 & 2021-09-28       & Large viewpoint variation         \\ \hline
			LH2 & 2021-09-28       & Different time period         \\ \hline
			LH3 & 2021-10-06       & Illumination variation            \\ \hline
			LH4 & 2021-10-11       & Pedestrian and partial occlusion \\ \hline
		\end{tabular}
	\end{center}
\end{table}

\begin{table}[h]
	\caption{Overview of the long hallway dataset}
	\label{table_example}
	\begin{center}
	  \begin{tabular}{cccc}
		\hline
		ID  & Date       & Time      & Environments                   \\ \hline
		LH1 & 2021-09-28 & Morning   & Large viewpoint variation         \\ \hline
		LH2 & 2021-09-28 & Evening   & Different time period         \\ \hline
		LH3 & 2021-10-06 & Afternoon & Illumination variation            \\ \hline
		LH4 & 2021-10-11 & Morning   & Pedestrian and partial occlusion \\ \hline
	  \end{tabular}
  \end{center}
\end{table}
\end{comment}

%----------------------------------------------------
\subsection{Mapping Performance}

In this section, we evaluate the performance of the proposed Object-level Semantic Topological Mapping method in terms of lightweight and on collected dataset.
Besides, we present the accuracy of our method qualitatively.

%\subsubsection{Lightweight}

\begin{figure}[htbp]
	\begin{center}
		\includegraphics[width=1\linewidth]{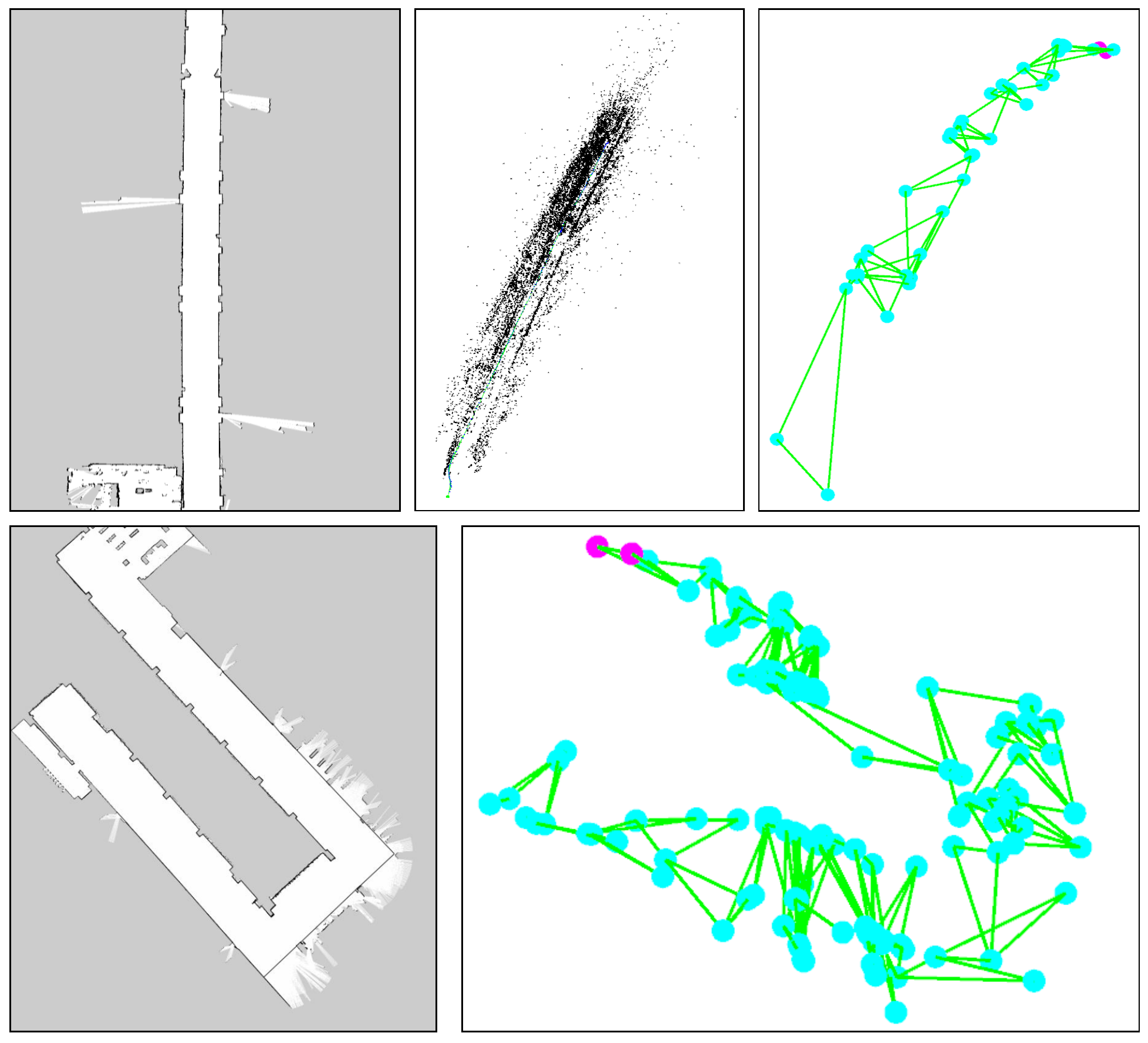}
	\end{center}
	\caption{Schematic representation of the three types of maps. The first row is a long indoor hallway environment with weak textures (In order, traditional occupied grid map, state-of-the-art sparse point cloud map and our map), and the second row is a large complex hospital scene (In order, traditional occupied grid map and our map).}
	\label{fig:9}
\end{figure}

Firstly, to demonstrate the lightweight performance of our mapping method, we compare our object-level semantic topological map with the typical grid map~\cite{c23} and a state-of-the-art point cloud map~\cite{c22}.
We perform experiments on both long hallway and hospital environment.
And we choose storage volume of the map as evaluation metric.
The experimental results are shown in Table \Rmnum{2} and Fig~\ref{fig:9}.
%It can be seen that, our map achieves more lightweight result than others in both long hallway and large scale environments.
As we can see from the table, in the long corridor scenario only, our map storage is nearly 23 times smaller than the grid map and 310 times smaller than the sparse point cloud map.
Thus, the experimental results show the more lightweight performance of the present method.
Besides, by comparing the amount of map storage for long corridors and hospital scenarios, we can observe that although the storage space of the map increases as the area increases, our approach greatly reduces the space of map storage and reduces the risk of storage overflow during long-term navigation due to the use of lightweight topology to build the map.
% to verify the lightweight performance of the topological semantic map, we compare the storage volume of the map with a grid map and a state-of-the-art point cloud map, as is shown in Table \Rmnum{2}.
% Since monocular initialization is very difficult in the test environment, therefore, the stereo configuration is used in the comparison experiments.
% However, stereo ORB-SLAM3 still cannot achieve effective mapping in hospital scenarios due to memory overflow and multiple losses, which also means that our approach has better lightweight and robustness.
% In addition, from the table, we can see that although the storage space of the map increases as the area increases, our approach greatly reduces the space of map storage and reduces the risk of storage overflow during long-term navigation due to the use of lightweight topology to build the map.
% %In particular, our method uses line features and achieves the best accuracy.
% Fig 7 shows the results of the three methods of mapping.
\begin{comment}
\begin{table}[h]
\caption{the storage volume of the map}
\label{table_example}
\begin{center}
\begin{tabular}{cccc}
\hline
& \textbf{Ours} & \textbf{Grid map} & \textbf{ORB-SLAM2} \\ \hline
Long hallway            & \textbf{66.4Kb}          & 1.52Mb  & 18Mb           \\ \hline
Hospital            & \textbf{233.7Kb}          & 3.51Mb  & --Mb           \\ \hline

\end{tabular}
\end{center}
\end{table}
\end{comment}
\begin{table}[h]
	\caption{the storage volume of the map}
	\label{table_example}
	\begin{center}
		\begin{tabular}{cccc}
			\hline
			& \textbf{Ours} & \textbf{Grid map} & \textbf{ORB-SLAM3} \\ \hline
			Long hallway            & \textbf{66.4Kb}          & 1.52Mb  & 20.6Mb           \\ \hline
			Hospital            & \textbf{233.7Kb}          & 3.51Mb  & --Mb           \\ \hline
			
		\end{tabular}
	\end{center}
\end{table}

Secondly, the time analysis of the main time-consuming parts of the whole system is shown in Table \Rmnum{3}.
The time consuming part of object detection contains the acquisition of semantic information about the object and the computation of 3D centroids. 
The speedup is approximately ten times compared to the traditional visual feature point extraction.
%As we can see from the table, our system can run in real time.
%The addition of semantic objects does not increase the computational burden of the system.
As we can see from the table, although the addition of semantic objects increases the computational burden of the system, our mapping system can still run online.
\begin{table}[h]
	\caption{Time-consuming for each module of our method}
	\label{table_example}
	\begin{center}
		\begin{tabular}{cc}
			\hline
			Module                                    & Time(ms/frame) \\ \hline
			Object Extraction                         & 89             \\ \hline
			\multicolumn{1}{l}{Descriptor Extraction} & 0.58           \\ \hline
			Graph Matching                            & 0.89           \\ \hline
			Total                                     & 90.47          \\ \hline
		\end{tabular}
	\end{center}
\end{table}
\begin{comment}
\begin{figure}[htbp]
	\begin{center}
		\includegraphics[width=1\linewidth]{10-eps-converted-to.pdf}
	\end{center}
	\caption{The complex hospital scenes.}
	%\label{fig:long}
	\label{fig:hospital}
\end{figure}
\end{comment}

\begin{figure}
	\begin{minipage}[t]{0.5\linewidth}
		\centering
		\includegraphics[scale=0.28]{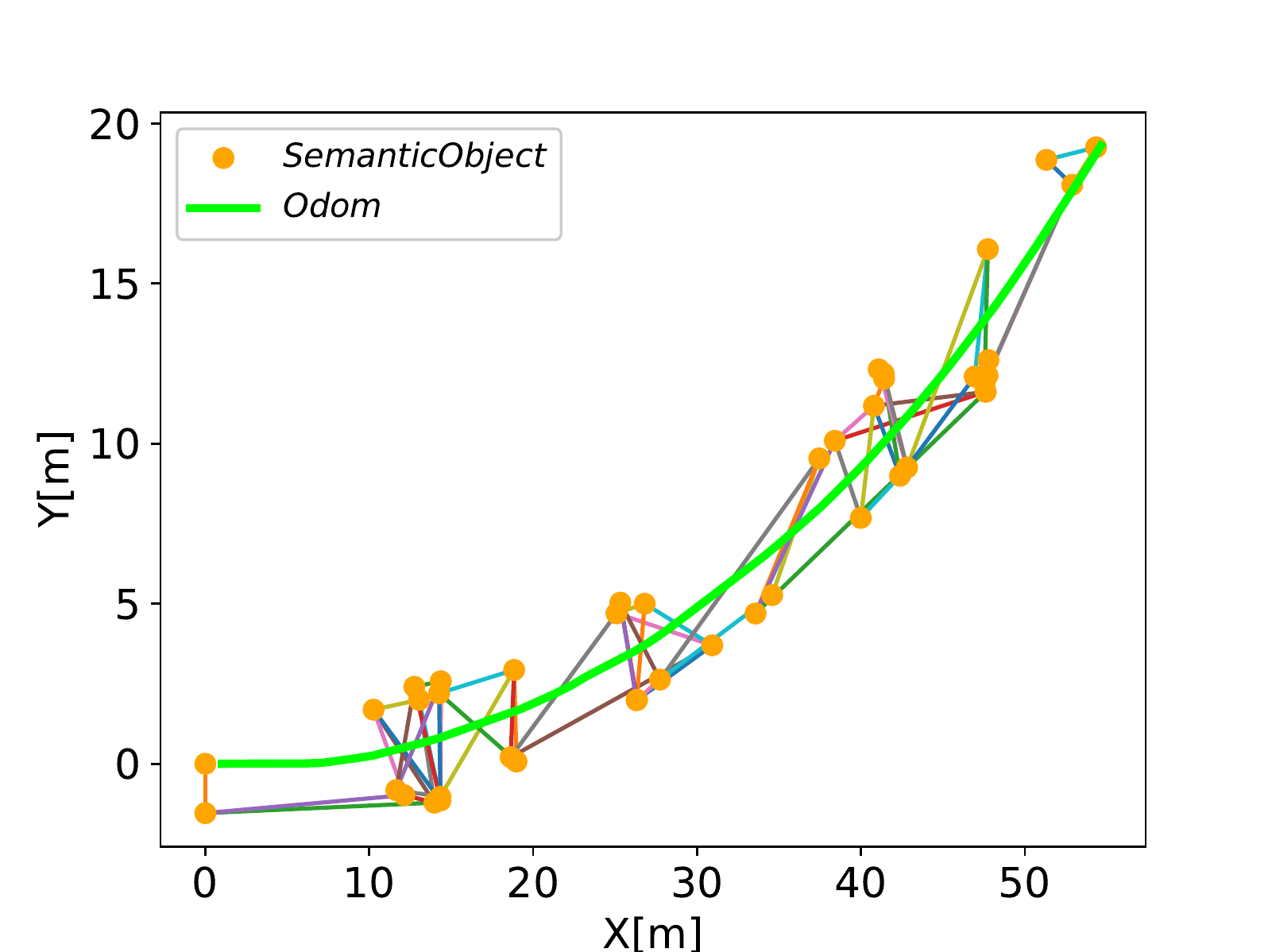}
		\centerline{(a) Long hallway scenes}
	\end{minipage}%
	\begin{minipage}[t]{0.5\linewidth}
		\centering
		\includegraphics[scale=0.28]{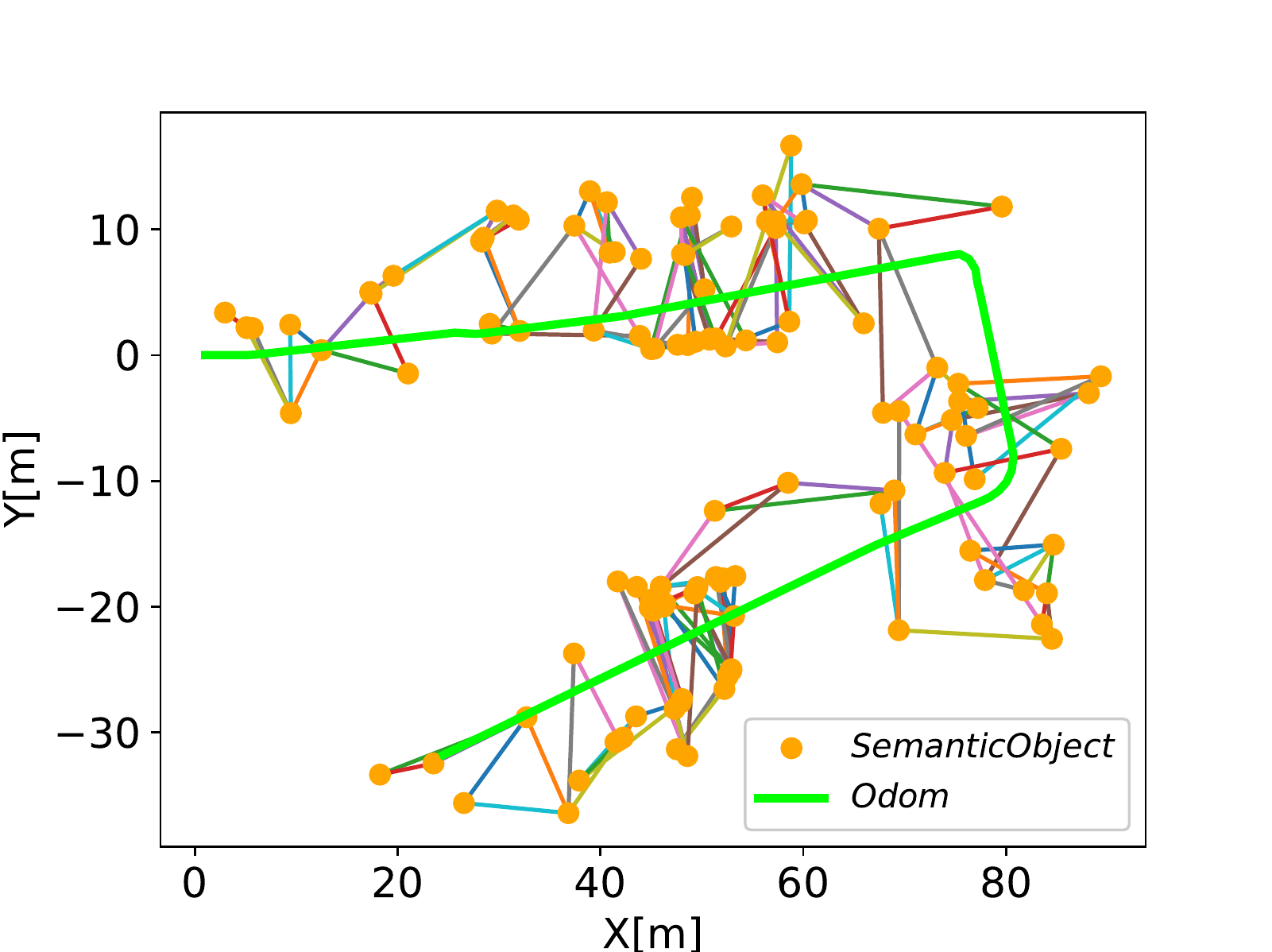}
		\centerline{(b) Hospital scenes}
	\end{minipage}
    \caption{The mapping proformance comparison experimental in the long hallway and hospital scenarios. The orange nodes are semantic objects and the colored line segments are edges between adjacent objects. The green curves are the odometer trajectories.}	
    \label{fig:10}	
\end{figure}

Finally, to further evaluate the accuracy of the maps, we mapped the constructed robot-centric topological semantic maps to the odometer coordinate system by coordinating transformation and compared them with wheeled odometers.
The experimental results in the long hallway and hospital scenarios are shown in Fig~\ref{fig:10}.
It can be seen from the figure that the constructed map basically matches the trajectory of the odometer, which verifies the accuracy of the map.
\begin{comment}
\begin{figure*}[]	
	%\begin{tabular}{cc}	
	\begin{minipage}{0.5\linewidth}		
		\centerline{\includegraphics[width=9.8cm]{9plothall-eps-converted-to.pdf}}		
		\centerline{(a) Long hallway scenes}		
	\end{minipage}	
	\hfill	
	\begin{minipage}{0.5\linewidth}		
		\centerline{\includegraphics[width=9.8cm]{9plothos-eps-converted-to.pdf}}		
		\centerline{(b) Hospital scenes}		
	\end{minipage}	
	\vfill		
	%\end{tabular}	
	\caption{The mapping proformance comparison experimental in the long hallway and hospital scenarios. The orange nodes are semantic objects and the colored line segments are edges between adjacent objects. The green curves are the odometer trajectories.}	
	\label{fig:map}	
\end{figure*}
\end{comment}
%\subsection{Navigation in real world (optional)}

\begin{comment}
\begin{table}[h]
\caption{An Example of a Table}
\label{table_example}
\begin{center}
\begin{tabular}{|c||c|}
\hline
One & Two\\
\hline
Three & Four\\
\hline
\end{tabular}
\end{center}
\end{table}
\end{comment}

\begin{comment}
\begin{figure}[thpb]
\centering
\framebox{\parbox{3in}{We suggest that you use a text box to insert a graphic (which is ideally a 300 dpi TIFF or EPS file, with all fonts embedded) because, in an document, this method is somewhat more stable than directly inserting a picture.
}}
%\includegraphics[scale=1.0]{figurefile}
\caption{Inductance of oscillation winding on amorphous
magnetic core versus DC bias magnetic field}
\label{figurelabel}
\end{figure}

\end{comment}

%-------------------------------------------------
\subsection{Long-term Localization Performance}

In this section, to better identify the strengths and weakness of our semantic scene graph matching based localization method, we comparing it with several leading appearance-based localization methods in the collected long-term dynamic scenarios.
% conduct experiments from four common complex environmental changes caused by temporal and spatial variations. 
% In this section, to fully and effectively evaluate the performance of long-term localization methods based on semantic scene graph matching by comparing it with several leading  appearance-based localization methods in typical long-term dynamic scenarios.
% The details are described as follows.

\begin{figure}[htbp]
	\begin{center}
		\includegraphics[width=1\linewidth]{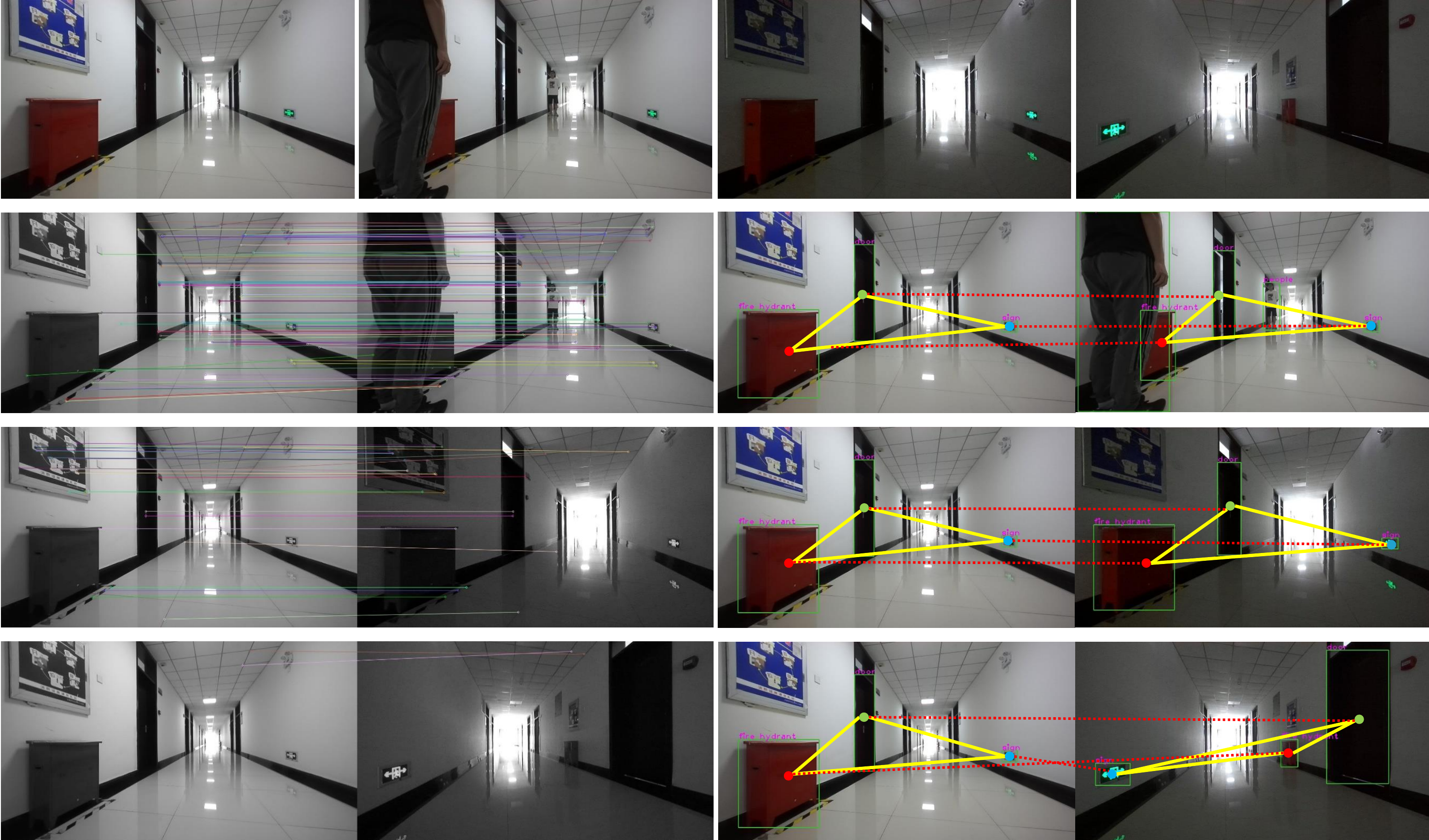}
	\end{center}
	\caption{Three examples of challenging scenarios (large viewpoint difference, illumination change and dynamic object) in localization. First row: query frame and localization frame
		taken in the same place. Three columns in the lower left: Results of detection based on ORB features. Three columns at the bottom right: Results of detection based on our semantic object.}
	\label{fig:11}
	%\label{fig:onecol}
\end{figure}

To illustrate the advantages of the our algorithm, firstly, we performed two feature matching experiments in three challenging scenes (viewpoint change, lighting change, and dynamic objects), as is shown in Fig~\ref{fig:11}. 
The three columns in the lower left corner are the detection results based on the traditional ORB features. 
From the figure, we can see that the accuracy of the method is susceptible to environmental changes and cannot effectively perform scene similarity matching. 
However, we can observe that the semantic mapping matching-based approach has better robustness to larger differences in environmental changes.

\begin{figure*}[ht]	
	%\begin{tabular}{cc}	
	\begin{minipage}{0.5\linewidth}		
		\centerline{\includegraphics[width=9.8cm]{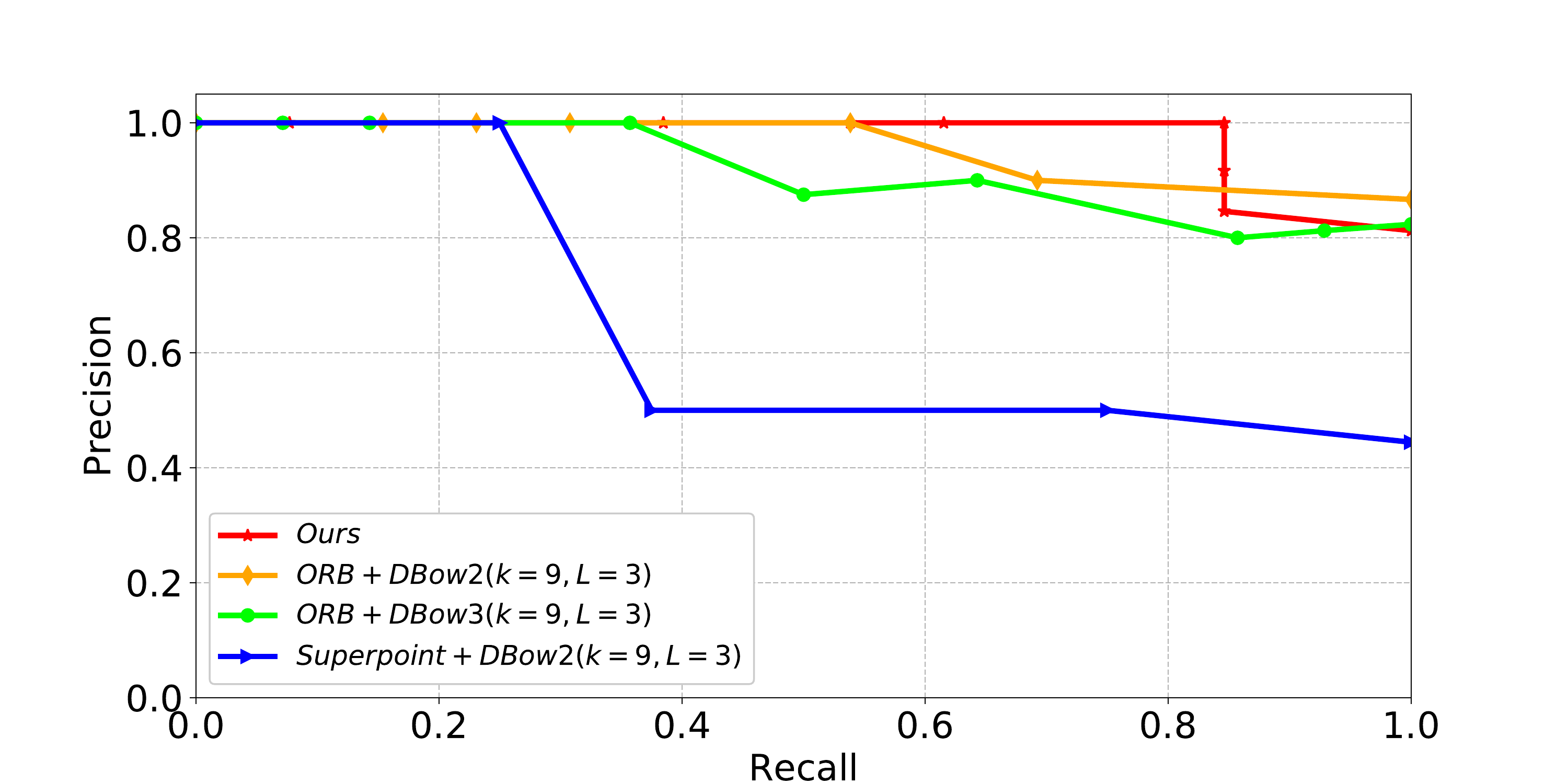}}		
		\centerline{(a) LH0-LH2, Day and night time periods}		
	\end{minipage}	
	\hfill	
	\begin{minipage}{0.5\linewidth}		
		\centerline{\includegraphics[width=9.8cm]{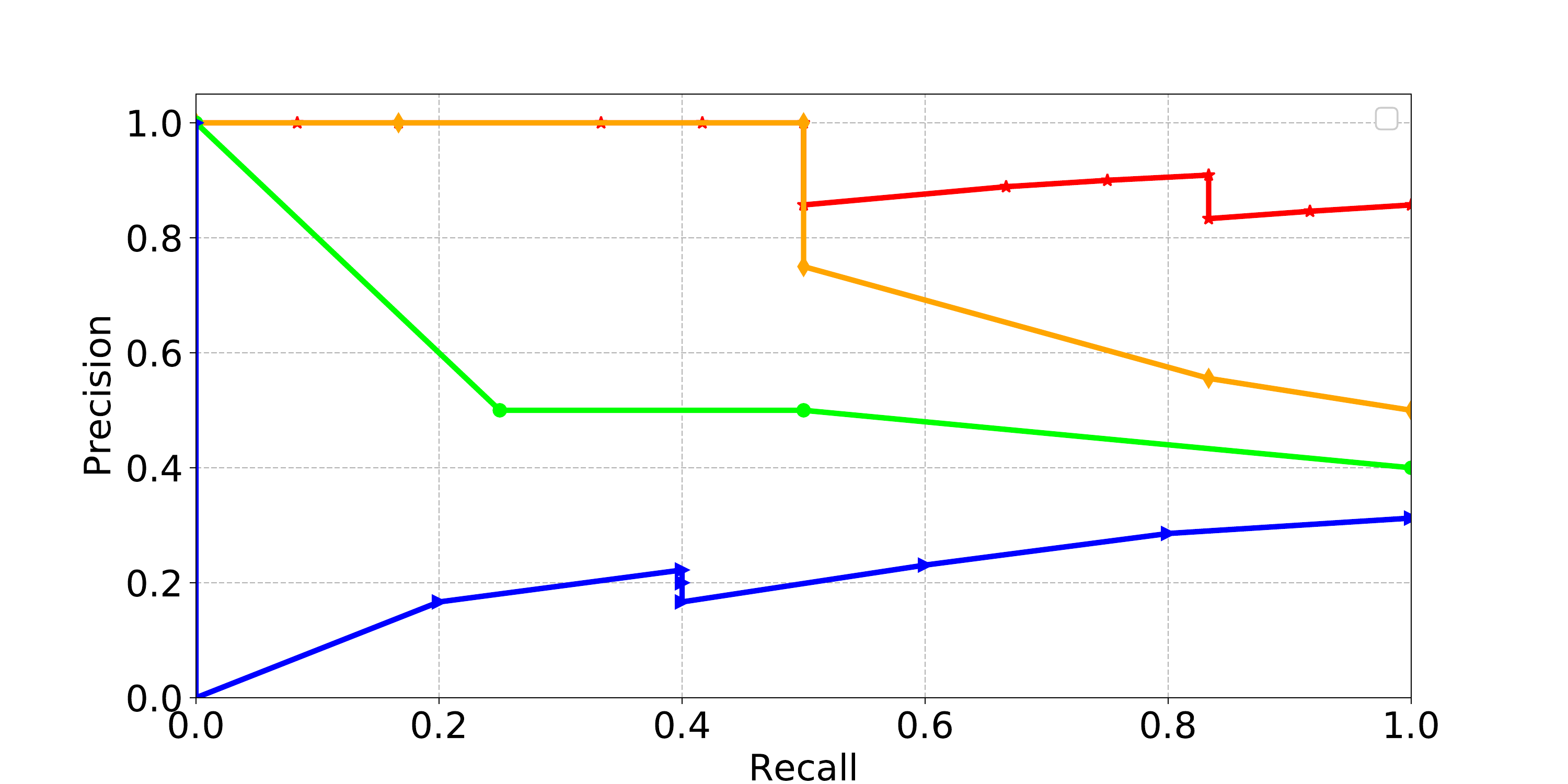}}		
		\centerline{(b) LH0-LH3, Illumination variation}		
	\end{minipage}	
	\vfill	
	\begin{minipage}{0.5\linewidth}		
		\centerline{\includegraphics[width=9.8cm]{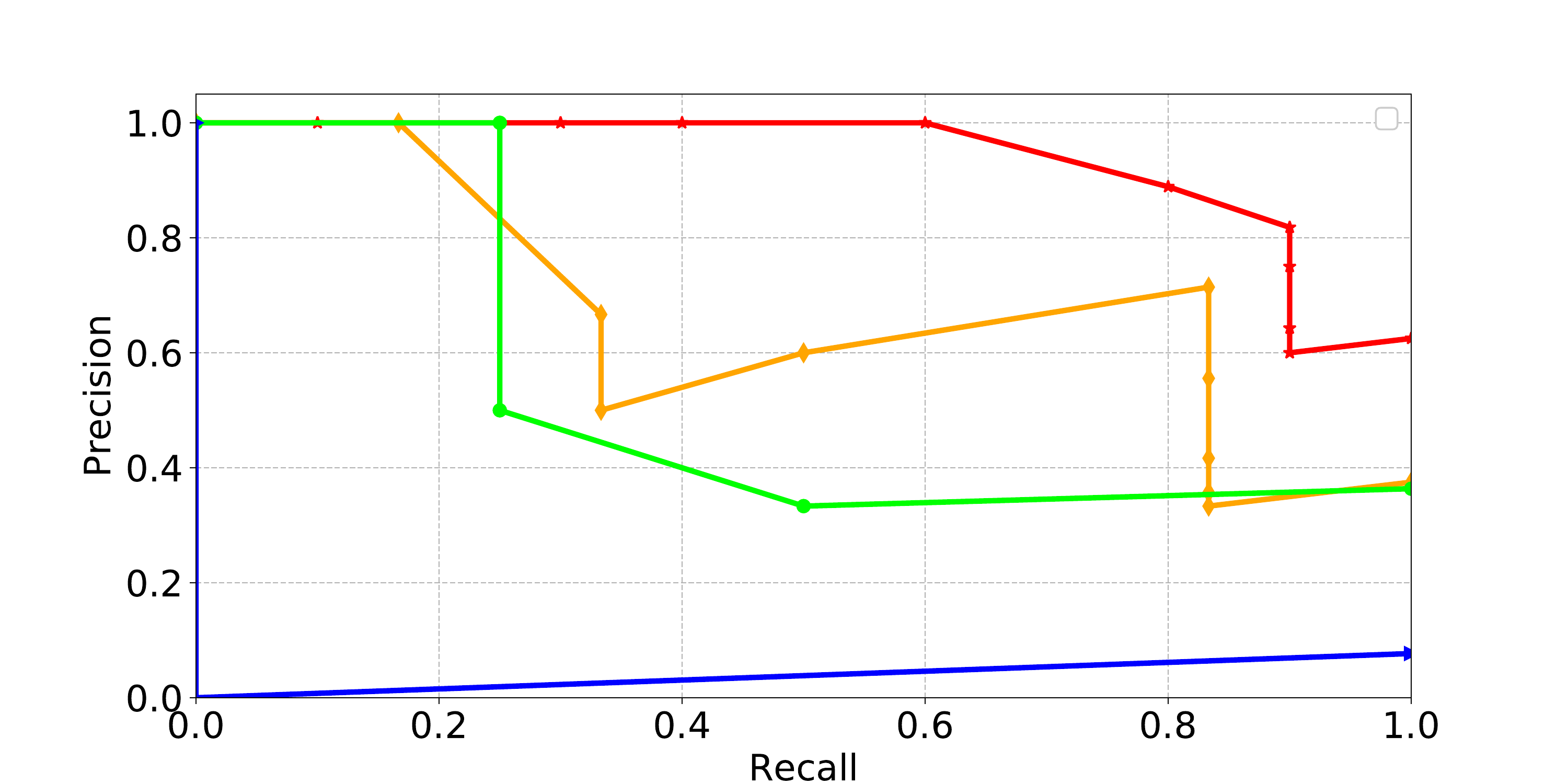}}		
		\centerline{(c) LH0-LH4, Pedestrian and partial occlusion}		
	\end{minipage}	
	\hfill	
	\begin{minipage}{0.5\linewidth}		
		\centerline{\includegraphics[width=9.8cm]{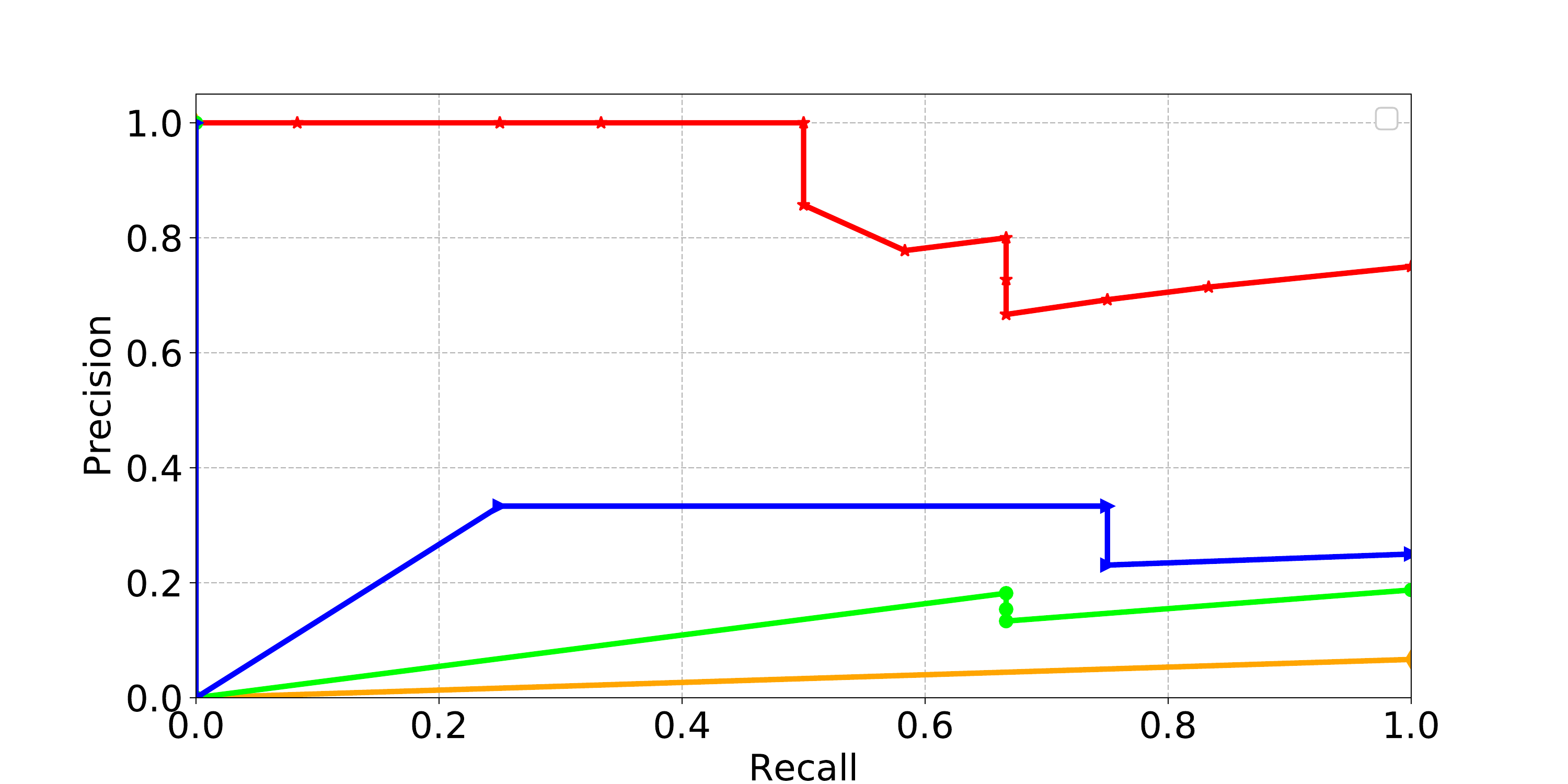}}	
		\centerline{(d) LH0-LH1, Large viewpoint variation}		
	\end{minipage}	
	\vfill	
	%\end{tabular}	
	\caption{PR curves for localization experiments based on long hallway datasets. (a) illustrates the effect of different time periods settings on the localization performance. (b) shows the effect of changing the illumination, while (c) depicts the effect of using dynamic object and occlusion. In (d) we evaluated the impact of localization performance under large spatial viewpoint variations. Several appearance-based methods are taken as baseline methods, such as ORB-based BoW~\cite{c15} and Superpoint-based~\cite{c16}. K,L is the branch factor and depth level, respectively.}	
	\label{fig:13}	
\end{figure*}
Secondly, we compared it with several appearance-based methods. 
The first one is a bag of words (BoW) approach using traditional ORB features, built on DBow2\footnote{https://github.com/dorian3d/DBoW2} and DBow3\footnote{https://github.com/rmsalinas/DBow3} libraries, respectively.
The second one is a BoW approach using CNN-based features extraction and description algorithm called Superpoint~\cite{c16}.
In the experiment, Precision-Recall curves (PR-curve) are used to represent the performance of localization. The experimental results are shown in Fig~\ref{fig:13}.
Fig~\ref{fig:13}.a first shows that several methods have similar accuracy for different time periods only, due to the fact that the less variation of light in the closed indoor environment.
%Fig 11.a first shows that several methods have similar accuracy at different time periods dataset, due to the fact that the less variation of light in the closed indoor environment.
%This is mainly due to the fact that the localization is less subject to visible light variations in the indoor closed environment. 
However, the accuracy of the appearance-based approach significantly decreased when the lights in the corridor are turned off resulting in changes in illumination, as is shown in Fig~\ref{fig:13}.b.
Particularly, Fig~\ref{fig:13}.c presents the effect of dynamic objects and partial occlusion on localization, and the results show better robustness of the semantic scene graph matching-based approach, due to the enhanced tolerance to occlusion by the local scene graph descriptors of the objects.
What’s more, Fig~\ref{fig:13}.d further verifies that our method has a significant advantage over appearance-based methods under viewpoint variation.
As can be seen from all the subplots, our methods not only have high accuracy, but also have a better recall rate than these appearance-based methods.
In addition, even CNN-based methods can be difficult to achieve effective feature extraction and description in difficult structured environments with sparse features and textures.
Meanwhile, to show the comparison results of the P-R curves more clearly, we calculated the AUC value of each scene, as is shown in Table \Rmnum{4}. It proves the robustness of the proposed method
further.
\begin{table}[htbp]
	\caption{the AUC values for each environment}
	\label{table_example}
	\begin{center}
	  \begin{tabular}{ccccc}
	  	\hline
	  	ID  & Ours          & ORB\_DBow2 & ORB\_DBow3 & SUP\_DBow2 \\ \hline
	  	LH1 & \textbf{0.59} & 0.13       & 0.28       & 0.57       \\ \hline
	  	LH2 & \textbf{0.92} & 0.89       & 0.73       & 0.61       \\ \hline
	  	LH3 & \textbf{0.83} & 0.83       & 0.79       & 0.25       \\ \hline
	  	LH4 & \textbf{0.86} & 0.73       & 0.72       & 0.37       \\ \hline
	  \end{tabular}
    \end{center}
\end{table}
\begin{figure}[htbp]
	\begin{center}
		\includegraphics[width=1\linewidth]{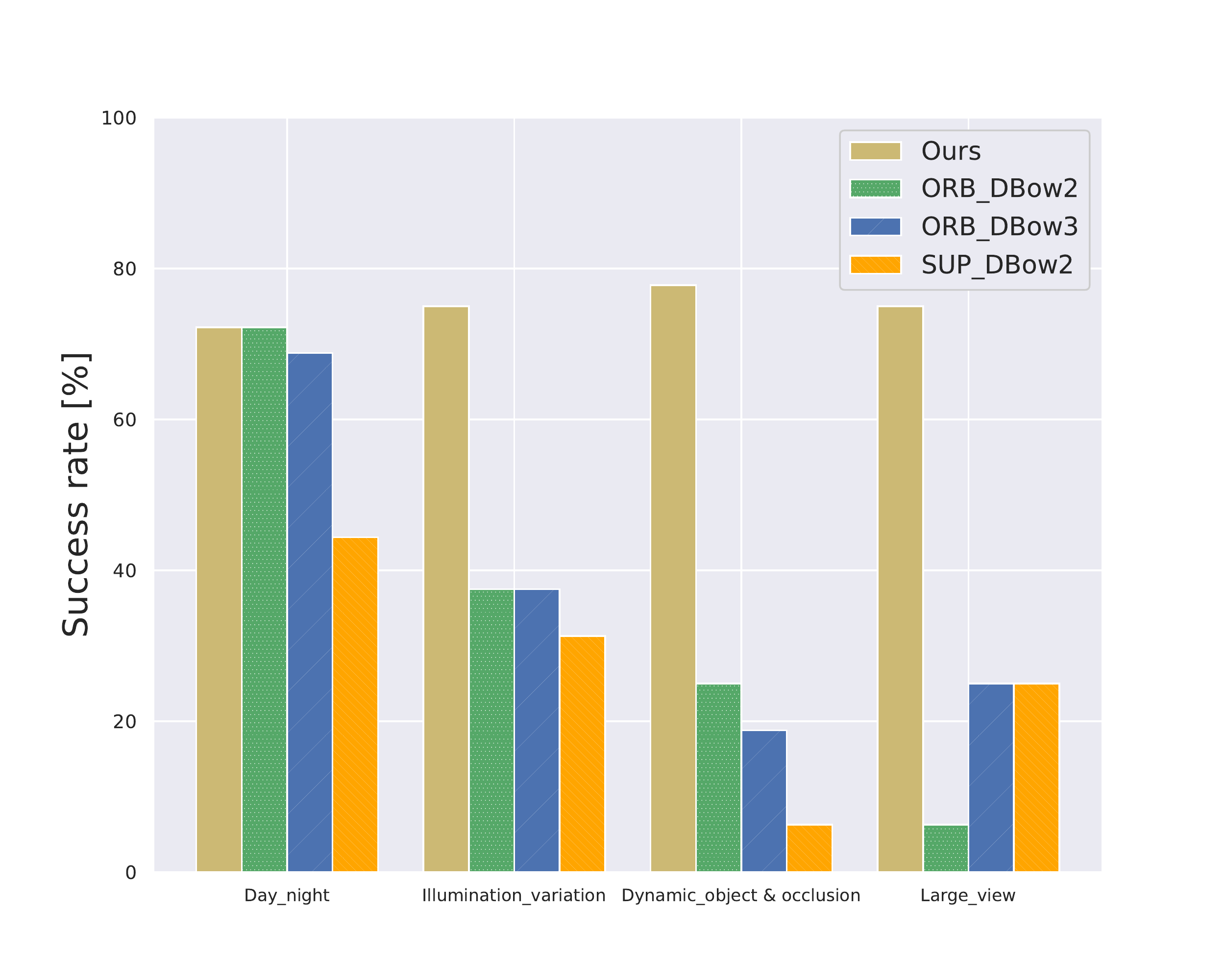}
	\end{center}
	\caption{The success rate of localization.}
	\label{fig:12}
	%\label{fig:onecol}
\end{figure}

Last but not least, we statistically depicted the success rate of localization in Fig~\ref{fig:12}.
Our method has a higher success rate than appearance-based techniques on multiple datasets and stays close to the 80th percentile.

\section{CONCLUSIONS}

In this letter, we present a novel lightweight and robust object-level topological semantic mapping and localization framework for long-term navigation.
On the one hand, we constructed a lightweight topological semntic map with defined learning topological primitives and hierarchical memory management mechanisms.
On the other hand, we achieve robust localization based on improved semantic scene graph descriptors of object and graph matching methods.
The framework is evaluated on challenging weakly textured long hallway and large scale hospital scenario datasets, and was run on a low cost embedded computing platform.
%Our approach demonstrates both high accuracy and robustness against drastic scene and viewpoint variations where others will be seriously affected.
Experimental results show that the method is sufficiently lightweight and highly robust at large scales, unstructured and long-term dynamic situations and with limited computational resources.
In the future we will continue to optimize the performance of mapping and localization, and explore the utilization of constructed maps to perform long-term navigation tasks in real-world scenarios.

\addtolength{\textheight}{-12cm}   % This command serves to balance the column lengths

\end{document}